\documentclass[acmlarge,authorversion,nonacm]{acmart}

\makeatletter
\def\@ACM@checkaffil{
    \if@ACM@instpresent\else
    \ClassWarningNoLine{\@classname}{No institution present for an affiliation}%
    \fi
    \if@ACM@citypresent\else
    \ClassWarningNoLine{\@classname}{No city present for an affiliation}%
    \fi
    \if@ACM@countrypresent\else
        \ClassWarningNoLine{\@classname}{No country present for an affiliation}%
    \fi
}
\makeatother

\usepackage{amsmath,amsfonts,bm}

\def\eqref#1{equation~\ref{#1}}

\def\1{\bm{1}}

\DeclareMathAlphabet{\mathsfit}{\encodingdefault}{\sfdefault}{m}{sl}
\SetMathAlphabet{\mathsfit}{bold}{\encodingdefault}{\sfdefault}{bx}{n}

\newcommand{\msg}[3]{m_{#1\rightarrow #2}^{#3}(X_{#2})}
\newcommand{\bel}[2]{bel_{#1}^{#2}(X_{#1})}
\newcommand{\prt}[3]{\mu_{#1#2}^{(#3)}}
\newcommand{\wgt}[3]{w_{#1#2}^{(#3)}}
\newcommand{\set}[1]{\{#1\}}

\newcommand{\unary}[3]{%
\ifthenelse{\equal{#2}{}}{\phi_{#1}(X_{#1},Y_{#1})}{\ifthenelse{\equal{#3}{}}{\phi_{#1}(X_{#1}=#2,Y_{#1})}{\phi_{#1}(X_{#1}=#2,Y_{#1}=#3)}}%
}
\newcommand{\pairwise}[4]{%
\ifthenelse{\equal{#3}{}}
{\psi_{#1#2}(X_{#1},X_{#2})}
{\ifthenelse{\equal{#4}{}}
  {\psi_{#1#2}(X_{#1},X_{#2}=#3)}
  {\psi_{#1#2}(X_{#1}=#3,X_{#2}=#4)}}%
}

\newcommand{\temporal}[1]{\tau_{#1}(\epsilon)}

\usepackage{hyperref}
\hypersetup{
    colorlinks,
    linkcolor={black},
    citecolor={black},
    urlcolor={blue!80!black}
}
\usepackage{url}
\usepackage{amsmath,amsfonts}
\usepackage{algorithmic}
\usepackage{graphicx}
\usepackage{textcomp}
\usepackage{xcolor}
\usepackage{subcaption}
\usepackage{xargs}
\usepackage{float}
\usepackage[capitalise]{cleveref}
\usepackage[ruled,vlined,linesnumbered]{algorithm2e}
\usepackage{wrapfig}

\definecolor{backgrnd}{RGB}{201,231,255}

\usepackage{soul}
\sethlcolor{backgrnd}

\AtBeginDocument{%
  \providecommand\BibTeX{{%
    \normalfont B\kern-0.5em{\scshape i\kern-0.25em b}\kern-0.8em\TeX}}}

\begin{document}

\title{DNBP: Differentiable Nonparametric Belief Propagation}

  
\author{Anthony Opipari}
\email{topipari@umich.edu}
\affiliation{%
  \institution{University of Michigan}
}

\author{Jana Pavlasek}
\email{pavlasek@umich.edu}
\affiliation{%
  \institution{University of Michigan}
}

\author{Chao Chen}
\email{joecc@umich.edu}
\affiliation{%
  \institution{University of Michigan}
}

\author{Shoutian Wang}
\email{shoutian@umich.edu}
\affiliation{%
  \institution{University of Michigan}
}

\author{Karthik Desingh}
\email{kdesingh@cs.washington.edu}
\affiliation{%
  \institution{University of Washington}
}

\author{Odest Chadwicke Jenkins}
\email{ocj@umich.edu}
\affiliation{%
  \institution{University of Michigan}
}

\renewcommand{\shortauthors}{Opipari, et al.}

\begin{abstract}
We present a differentiable approach to learn the probabilistic factors used for inference by a nonparametric belief propagation algorithm. Existing nonparametric belief propagation methods rely on domain-specific features encoded in the probabilistic factors of a graphical model. In this work, we replace each crafted factor with a differentiable neural network enabling the factors to be learned using an efficient optimization routine from labeled data. By combining differentiable neural networks with an efficient belief propagation algorithm, our method learns to maintain a set of marginal posterior samples using end-to-end training.
We evaluate our differentiable nonparametric belief propagation (DNBP) method on a set of articulated pose tracking tasks and compare performance with learned baselines. Results from these experiments demonstrate the effectiveness of using learned factors for tracking and suggest the practical advantage over hand-crafted approaches. The project webpage is available at: \href{https://progress.eecs.umich.edu/projects/dnbp/}{https://progress.eecs.umich.edu/projects/dnbp/}.
\end{abstract}



\maketitle


\section{Introduction}

Perceiving the pose of objects in space is a critical capability for robots operating in human environments.
Perception and tracking of articulated objects, such as kitchen tools and human figures, is particularly challenging due to their high-dimensional and continuous state spaces where self occlusions are ubiquitous.
Probabilistic graphical model inference is an approach with potential to compute articulated pose under these conditions of uncertainty. 
Nonparametric belief propagation (NBP) algorithms~\citep{nbp:SudderthIFW03,isard2003pampas} are a form of generative probabilistic inference that have proven effective for inference in visual perception tasks such as human pose tracking~\citep{looselimbed:SigalBRBI04} and articulated object tracking in robotic perception~\citep{bpposeest:DesinghLOJ19, pavlasek2020parts}. 
In addition to accounting for uncertainty in partially observable environments, these algorithms show promising computational properties in practice~\citep{bpposeest:DesinghLOJ19,Ortiz2021visualGBP,liu2022hardware}. 
By relying on local message passing, NBP algorithms are ammenable for implementation on distributed, heterogeneous computing platforms.

The adaptability of NBP algorithms to new applications, however, is limited by the need to define hand-crafted functions that describe the distinct statistical relationships in a particular dataset. 
Current methods that utilize NBP rely on extensive domain knowledge to parameterize these relationships.
Reducing the domain knowledge required by NBP methods would enable their use in a broader range of applications. 

\begin{figure}[t]
\begin{center}
\includegraphics[width=\linewidth,keepaspectratio]{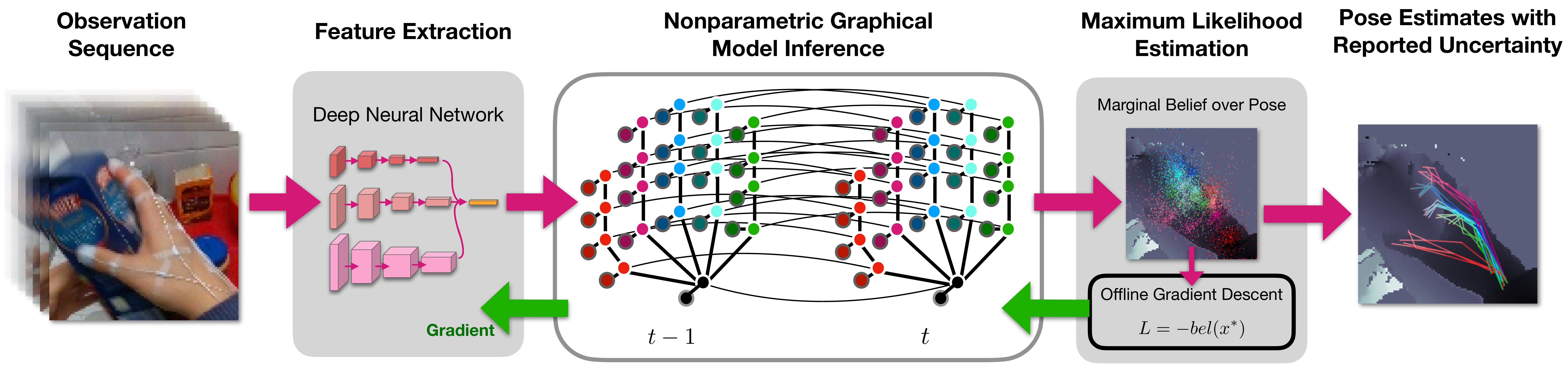}
\end{center}
\caption{\footnotesize Architecture diagram of differentiable nonparametric belief propagation. 
DNBP combines domain knowledge in the form of graphical models with differentiable neural networks for tractable inference in continuous spaces. 
Input features from a deep neural network and the probabilistic relationships encoded in a graphical model are learned jointly in an end-to-end fashion using backpropagation. 
Following offline training, DNBP can be applied to unseen data without hand-tuning.}
\label{fig:teaser}
\end{figure}

The capacity of NBP algorithms to perform inference using arbitrary graphs sets them apart from other generative inference algorithms such as the recursive Bayes filter~\citep{probrob:thrun} (e.g. particle filter~\citep{partfiltreview}) and has been shown to be important in computational perception because it allows for modeling of non-causal relationships~\citep{nbp:SudderthIFW03}. 
Neural network-based approaches are an alternative for computational perception~\citep{lecun2015deep,AlexNet,DanNet,xiang2018posecnn, dope:tremblay2018corl}. 
These methods generally avoid the need for extensive domain knowledge by learning from large amounts of labelled data. 
Data-driven approaches, however, are prone to noisy estimates and have limited capacity to represent uncertainty inherent in their output.
In robotic applications, both of these limitations negatively impact the ability for a robot to operate effectively in unstructured environments.

In this paper, we present a differentiable nonparametric belief propagation (DNBP) method, a hybrid approach which leverages neural networks to parameterize the NBP algorithm. 
Inspired by the differentiable particle filter from \citet{dpf:JonschkowskiRB18} and the pull message passing for nonparametric belief propagation (PMPNBP) algorithm~\citep{bpposeest:DesinghLOJ19}, we develop a differentiable nonparametric belief propagation algorithm. DNBP performs end-to-end learning of each probabilistic factor required for graphical model inference.  

The effectiveness of DNBP is demonstrated on two simulated articulated tracking tasks and on a real-world hand pose tracking task in challenging, noisy environments. 
An analysis of the learned probabilistic factors and resulting tracking performance is used to validate the approach. 
Results show that our approach can leverage the graph structure to report uncertainty about its estimates while significantly reducing the need for prior domain knowledge required by previous NBP methods. 
DNBP performs competitively in comparison to traditional learning-based approaches on the tracking tasks. 
Collectively, these results indicate that DNBP has the potential to be successfully applied to robotic perception tasks, where maintaining a notion of uncertainty throughout the inference is beneficial.

\subsection{Motivation and Societal Considerations}

In this work, we seek methods that combine the robustness of generative probabilistic inference with the speed, recall power, and general adaptability of discriminative neural networks.
Our aim is to find new {\it generative-discriminative} inference methods that can achieve the best of both worlds.

Our work on DNBP is motivated by a simple question: how can AI models be relied on in the face of uncertainty?
As roboticists, we experience first-hand the uncertainty inherent in our interactions with the physical world, and the errors which result from it.
It is not a matter of {\it if} mistakes occur during inference, but {\it when} mistakes will occur and {\it how} our systems can recover from such errors.
There are profound questions as to {\it who} will be impacted when mistakes occur, and what costs will be imposed by these mistakes.
\textit{In other words, even if a neural network is accurate 99\% of the time, how useful will it be if we do not know the 1\% of time it is wrong?}

A notable example of the cost of mistakes in discriminative models is the 2020 wrongful arrest and detention of Mr. Robert Williams arising from a false positive facial recognition output by a neural network~\cite{WashingtonPostOpEd}.
The potential for such harmful mistakes were foreseen in research from the algorithmic fairness community, such as work by~\citet{raji2020savingface} and~\citet{buolamwini2018gender}.
More broadly, the past decade has seen a transformative proliferation of discriminative neural networks~\cite{waibel1989phoneme,AlexNet,DanNet} as well as extraordinary growth in the body of research focused on algorithmic fairness~\cite{dwork2012fairness,kleinberg2017fairrisk}.\footnote{In addition, the impact of artificial intelligence has also led to massive growth in the demand for computer science degrees~\cite{CRATaulbee2020} as the most lucrative pathway into the field of artificial intelligence.
These growth trends have, unfortunately, not been seen for the participation of historically underrepresented minorities in professional pathways into computing and AI.
This national trend is indicated by efforts such as the \href{https://cse-climate.engin.umich.edu/reports/climate-dei-reports/cse-climate-dei-report-19-20/}{Michigan Computer Science and Engineering Climate, Diversity, Equity, and Inclusion Report}, which reported 7.5\% representation of students from underrepresented minority groups among its 2,586 declared undergraduate majors.
}
In complement, the methods of inference must also improve if AI systems are to be useful for responsible decision making in an uncertain world.
We posit that enabling AI models to maintain distributions over uncertain possibilities through the generation and evaluation of diagnosable hypotheses will empower human users seeking to diagnose when and why the models are unreliable.


An encouraging example of how generative inference can be reliable for robust AI comes from robotics and the foundational task of autonomous map building and localization~\citep{probrob:thrun}.
Robots that use Bayes filtering for localization~\cite{Dellaertetal1999icra,fox2002nips} are able to overcome localization errors by recursively diagnosing which possible location for the robot best aligns with its observation of the environment.
This generative approach to inference alternates between {\it hypothesising} possible states of the world and {\it evaluating} these hypotheses according to their agreement (or disagreement) with observation.
Using this generative process enables robots to maintain probability mass over plausible states which are updated over time in response to perceived localization error.
In contrast, a discriminative neural network approach to localization fuses together hypothesis and evaluation into a single opaque forward pass.
The discriminative model implicitly reasons over the full state space.
As such, any interpretation or diagnosis of the output of the neural network requires gathering meaning from the inner workings of the network, which remains an open area of research~\cite{Selvaraju_2017_ICCV,BARREDOARRIETA202082}.
Therefore, we underscore a critical feature of generative probabilistic inference is the {\it diagnosability} of mistakes in its estimates.
In this work, we investigate a combined approach that explicitly generates hypothetical states and then weights them according to a learned neural network. 
With this approach, we aim to realize the best of both generative and discriminative approaches.





In addition to uncertainty in inference, we also know that decision making must often occur in situations dominated by partial observations.
Our ability to reason in the physical world often relies on our beliefs about unseen objects just as much or perhaps more than it relies on what is directly in our field of view at any given moment. 
Reasoning in belief space offers one path to address such uncertainty by maintaining and reasoning over distributions, which allows for recovery from mistakes and recursive refinement as new observations are collected.
However, purely probabilistic approaches to inference (arising from work by~\citet{kaelbling1998pomdp}) have proven prohibitively slow computationally up to this point.
New approaches to efficient generative probabilistic inference offer increasingly viable algorithms for object permanence~\cite{ZhenICRA2020} and belief space planning~\cite{AlphonsusIROS2021}, although it remains unclear if these methods will be tractable for meaningful use.
In this regard, we envision that robust robot systems will increasingly use generative-discriminative perception to dovetail with replanning algorithms~\cite{somani2013despot,yoon2007ff,caelan2020replan,Alphonsus2022eleph} for decision making.
With this vision in mind, the current work sets out to address the challenge of modelling perceptual uncertainty using a generative-discriminative approach which could be used by a downstream belief-space replanning system.
The replanning paradigm follows long-established practices used for recovery by mobile robots during autonomous navigation when localization errors occur~\cite{Dellaertetal1999icra,fox2002nips,biswas2013cobots}
We posit this approach to replanning will generalize across many scenarios where discriminative AI is now deployed, and ultimately lead to more accountable systems and responsible standards.




\begin{figure}[b]
\begin{center}
\includegraphics[width=\linewidth,keepaspectratio]{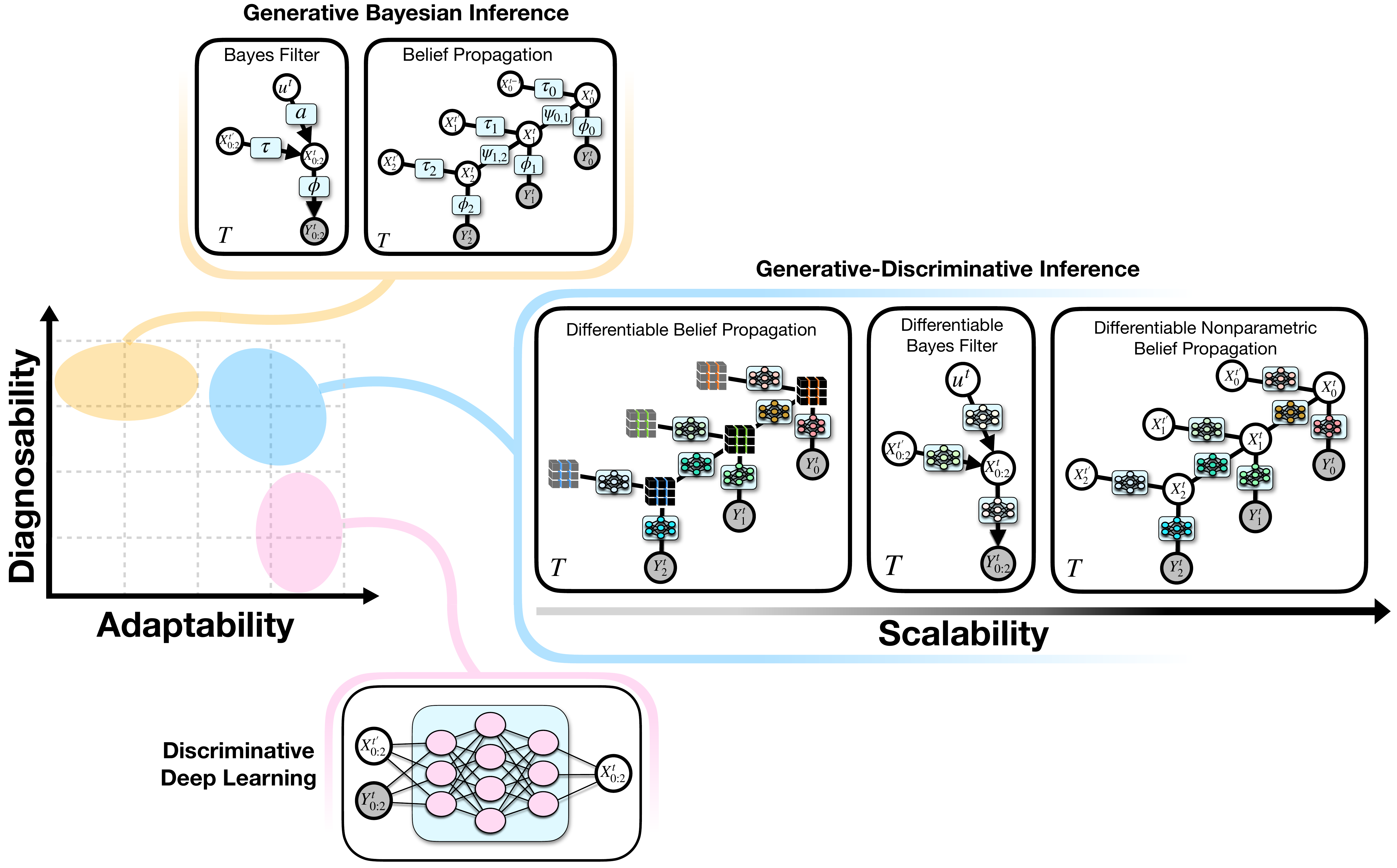}
\end{center}
\caption{\footnotesize A comparison of potential trade-offs among generative and discriminative inference approaches. Generative Bayesian inference (e.g. Bayes filtering~\citep{probrob:thrun} and belief propagation~\citep{bp:PEARL1988143}) exhibits a high degree of diagnosability but low adaptability stemming from their reliance on brittle hand-crafting~\citep{roy2021machine}. Discriminative deep learning~\citep{lecun2015deep} is a data-driven approach with high adaptability but can lacks diagnosability in dynamic, partially-observable and uncertain environments. The aim of this work is to investigate a hybrid generative-discriminative inference approach, Differential Nonparametric Belief Propagation (DBNP), that can achieve the best of both model-driven and data-driven techniques; it should have a high degree of diagnosability and adaptability. 
}
\label{fig:model_plot}
\end{figure}


\section{Related Work}

\textbf{Belief Propagation:}
In the context of graphical models, inference refers to the process by which information about observed variables is used to derive the posterior distribution(s) of unobserved random variables.
Belief propagation (BP) is a message passing algorithm for inferring the marginal distributions of graphical models~\citep{yedidia2003understanding}. BP computes exact marginal distributions on trees~\citep{bp:PEARL1988143}, and has demonstrated empirical success on graphs with cycles when applied in a loopy fashion~\citep{lbp:MurphyWJ99, bpstereo:SunZS03,bpface:LeeASG08,bpdenoise:LanRHB06}.
In order to apply inference techniques such as BP, the parameters of a graphical model (i.e. the graph structure and associated probabilistic factors) must be fully specified. 
The requirement that model parameters be specified limits BP's adaptability to new applications. 
Thus, BP is a model-driven approach to inference with a high degree of introspection but limited adaptability (top left of~\cref{fig:model_plot}).
Furthermore, BP demands exact integral computations that constrain the algorithm's applicability to state spaces that are discrete.
In contrast, this current study focuses on generative inference for continuous state spaces that robots are often faced with.

\noindent\textbf{Nonparametric Belief Propagation:}
For continuous spaces, such as six degrees-of-freedom object pose, exact integrals called for in BP become intractable and approximate methods for inference have been considered. Nonparametric belief propagation (NBP) methods \citep{isard2003pampas,nbp:SudderthIFW03}, have been proposed which represent the inferred marginal distributions using mixtures of Gaussians and define efficient message passing approximations for inference. 
\citet{isard2003pampas} demonstrated the effectiveness of their proposed algorithm using a set of synthetic visual datasets each modeled with hand-crafted factors. \citet{nbp:SudderthIFW03} applied their NBP method successfully to a visual parts-based face localization task as well as a human hand tracking task~\citep{sudderth2004visual}. 
In both applications, NBP relied on factor models which were chosen based on task-level domain knowledge (e.g. skin color statistics, valid hand configurations). \citet{looselimbed:SigalBRBI04} extended these NBP methods to human pose estimation and tracking using factors which were each trained apart from the inference algorithm using independent training objectives.

\citet{ihler2009particle} described a conceptual theory of particle belief  propagation, where messages being sent to inform the marginal of a particular variable could be generated using a shared proposal distribution. 
Following the work of Ihler and McAllester, \citet{bpposeest:DesinghLOJ19} presented an efficient ``pull'' message passing algorithm (PMPNBP) which uses a weighted particle set to approximate messages between random variables. 
PMPNBP is effective on robot pose estimation tasks using hand-crafted factors.
Using a similar approximation of belief propagation, \citet{pavlasek2020parts} took a step toward deep learning-based potential functions by introducing a pre-trained image segmentation network to the unary factors. 

An important limitation of the existing NBP methods is they assume the probabilistic factors expressed in the graph are provided as input or rely on domain knowledge to separately model and train each function.
However, the success of NBP methods in enabling inference that efficiently factors continuous state spaces motivates our work aimed at improving their adaptability.

\noindent\textbf{Deep Learning:} In recent years, neural network-driven deep learning has achieved state-of-the-art performance across a variety of perception tasks~\citep{lecun2015deep,redmon2016you,xiang2018posecnn,Guler_2018_CVPR}.
The ability for deep learning models to reliably estimate their uncertainty has been identified as an important challenge for applying these techniques to robotic domains~\citep{sunderhauf2018RobLims}.
Bayesian deep learning approaches have been developed to address this challenge for domains where quantified uncertainty and the potential for introspection is expected~\citep{abdar2021UQReview}.
Bayesian deep learning approaches for uncertainty quantification include Monte Carlo dropout~\citep{gal2016dropout,qendro2021Stoch}, variational inference~\citep{kingma2017variational,Caramalau_2021_WACV} and callibration~\citep{guo2017calib}.
The current study sets out to study a hybrid approach to enable uncertainty quantification by hybridizing deep neural networks with a nonparametric belief propagation algorithm.

\noindent\textbf{Differentiable Belief Propagation:} Deep learning architectures have been proposed that emulate the message passing operations of BP using tensor decompositions~\citep{dupty2020HighBP}, invertible neural operators~\citep{kuck2020BPNN}, convolutional neural networks~\citep{cnngraph:TompsonJLB14}, and graph neural networks~\citep{yoon2019GNNInf,zhang2020FGNN,satorras2021NeurBP}.
These hybrid approaches were found to outperform non-hybrid models on a variety of inference datasets. However, they are either limited to discrete spaces~\citep{dupty2020HighBP,kuck2020BPNN,cnngraph:TompsonJLB14,yoon2019GNNInf,satorras2021NeurBP} or provide only point estimates without a measure of uncertainty~\citep{zhang2020FGNN}.
\citet{XiongR20} proposed a variational inference approach to approximate BP using learned neural network potentials and Gaussian quadrature. The variational approach demonstrated promising inference performance on discrete classification and a synthetic trivariate Gaussian mixture dataset.
In contrast, the current work focuses on a NBP-deep learning hybrid for inference in continuous state spaces containing a larger number of degrees of freedom. 


\noindent\textbf{Differentiable Bayes Filtering:} Variations of the Bayes filtering algorithm have been applied successfully to continuous state space inference tasks in robotics~\citep{Dellaertetal1999icra,fox2002nips,probrob:thrun}. A new application of Bayesian filtering for robotics incorporates deep learning with end-to-end training. \citet{bpkf:HaarnojaALA16} introduced a differentiable Kalman filter for mobile robot state estimation.
\citet{dpf:JonschkowskiRB18} and \citet{pfnet:KarkusHL18} both proposed differentiable particle filter algorithms for modeling continuous state spaces. 
\citet{multimodfilt:abs-2010-13021} investigate how multimodal sensor information may be fused by deep learning and Bayesian filtering for rigid body pose estimation.
\citet{yi2021differentiable} propose an end-to-end learning method for inference over factor-graph models in tracking and localization tasks. 
These studies all model a single object body using variants of the Bayes filter. In contrast, the current study focuses on modeling multi-part, articulated objects within the robotic context.
The articulated object distinction motivates the use of NBP since its ability to factor high-dimensional continuous spaces is associated with improved performance in the face of increased dimensionality~\citep{bpposeest:DesinghLOJ19,pavlasek2020parts}.




\section{Belief Propagation}

\begin{figure}[t!]
\begin{center}
\begin{subfigure}[b]{3.95cm}
\includegraphics[height=3.95cm,keepaspectratio]{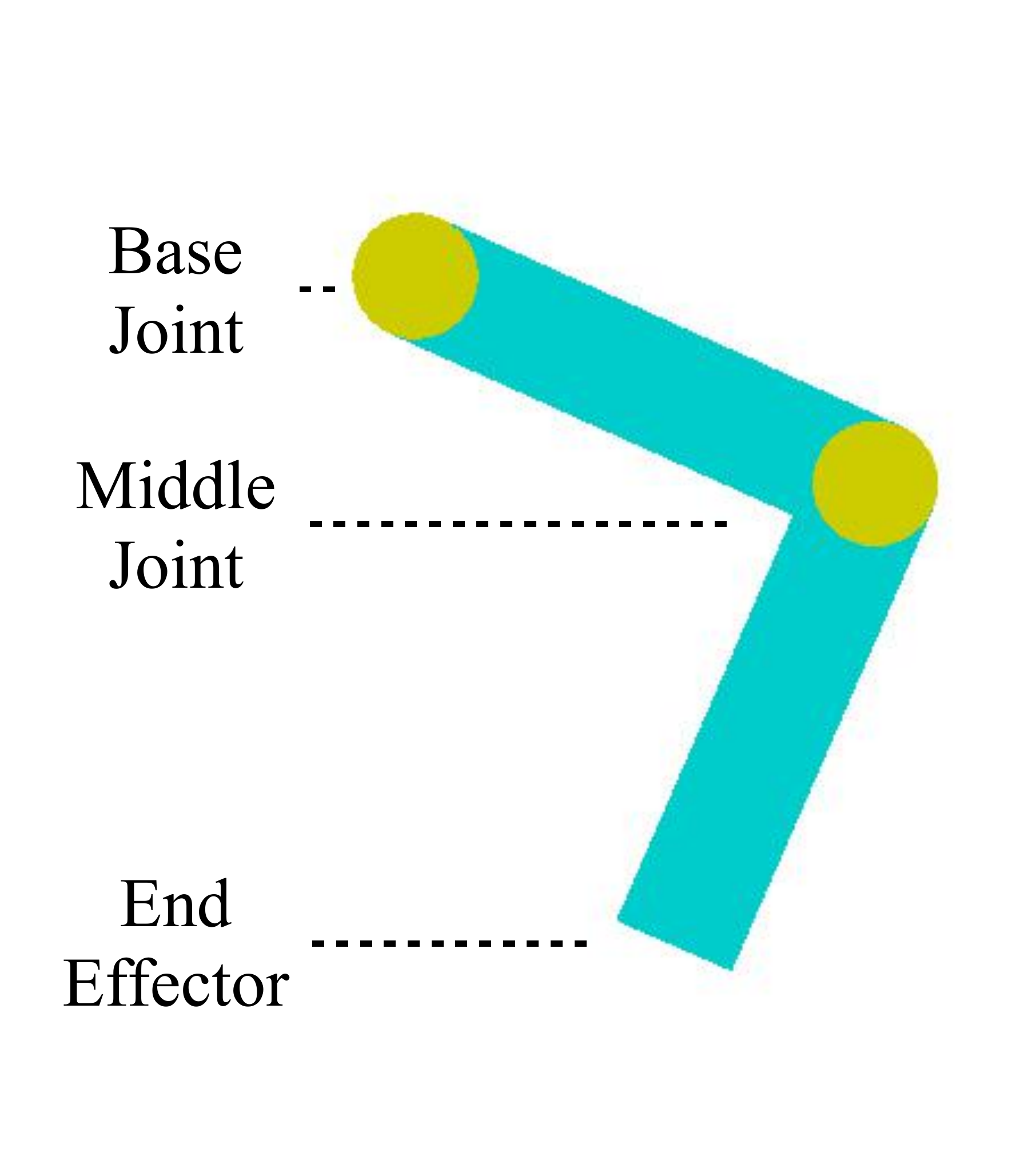}
\caption{}
\label{fig:pendulum_example}
\end{subfigure}\hfill%
\begin{subfigure}[b]{3.95cm}
\includegraphics[height=3.95cm,keepaspectratio]{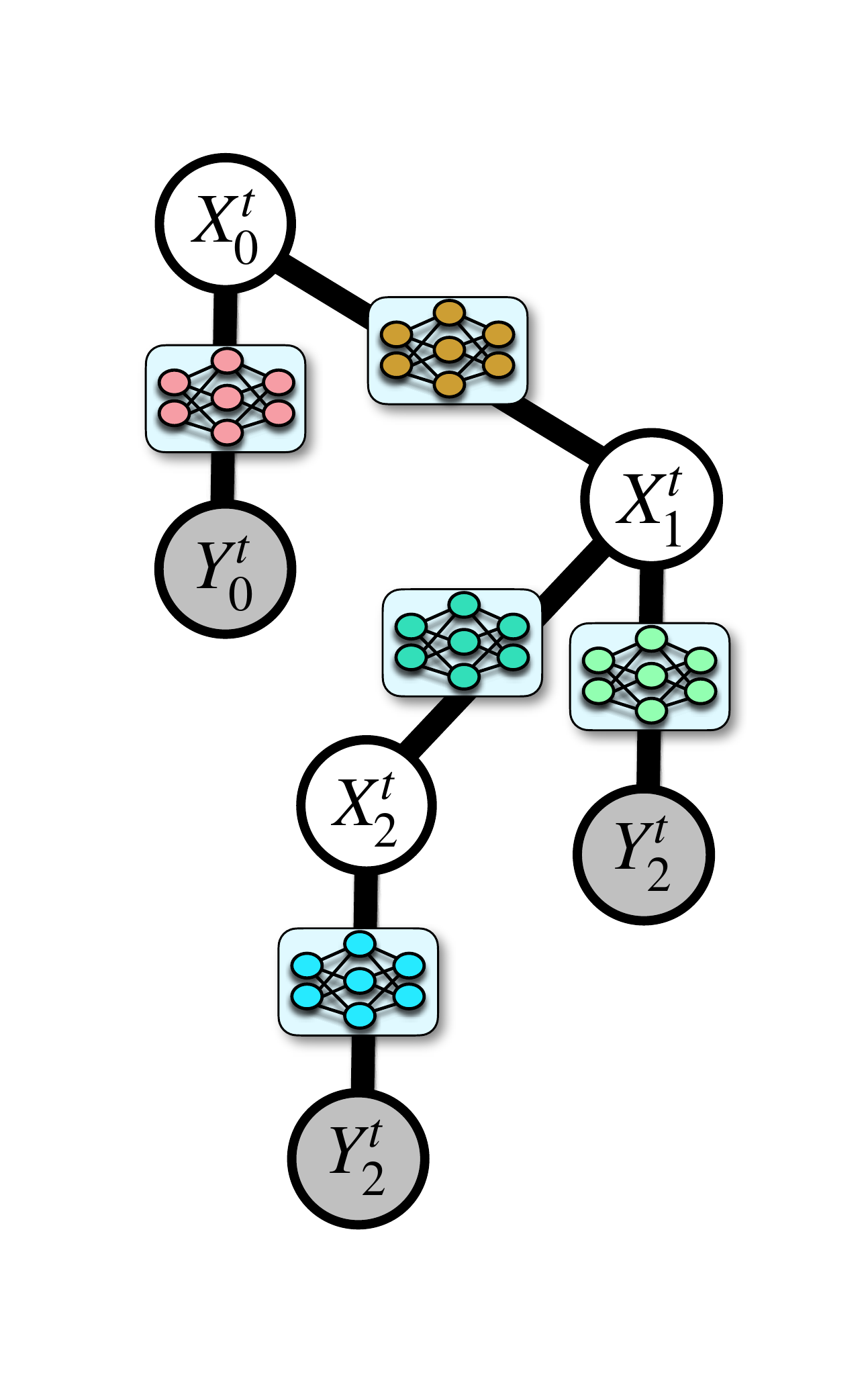}
\caption{}
\label{fig:pendulum_model}
\end{subfigure}
\begin{subfigure}[b]{3.95cm}
\includegraphics[height=3.95cm,keepaspectratio]{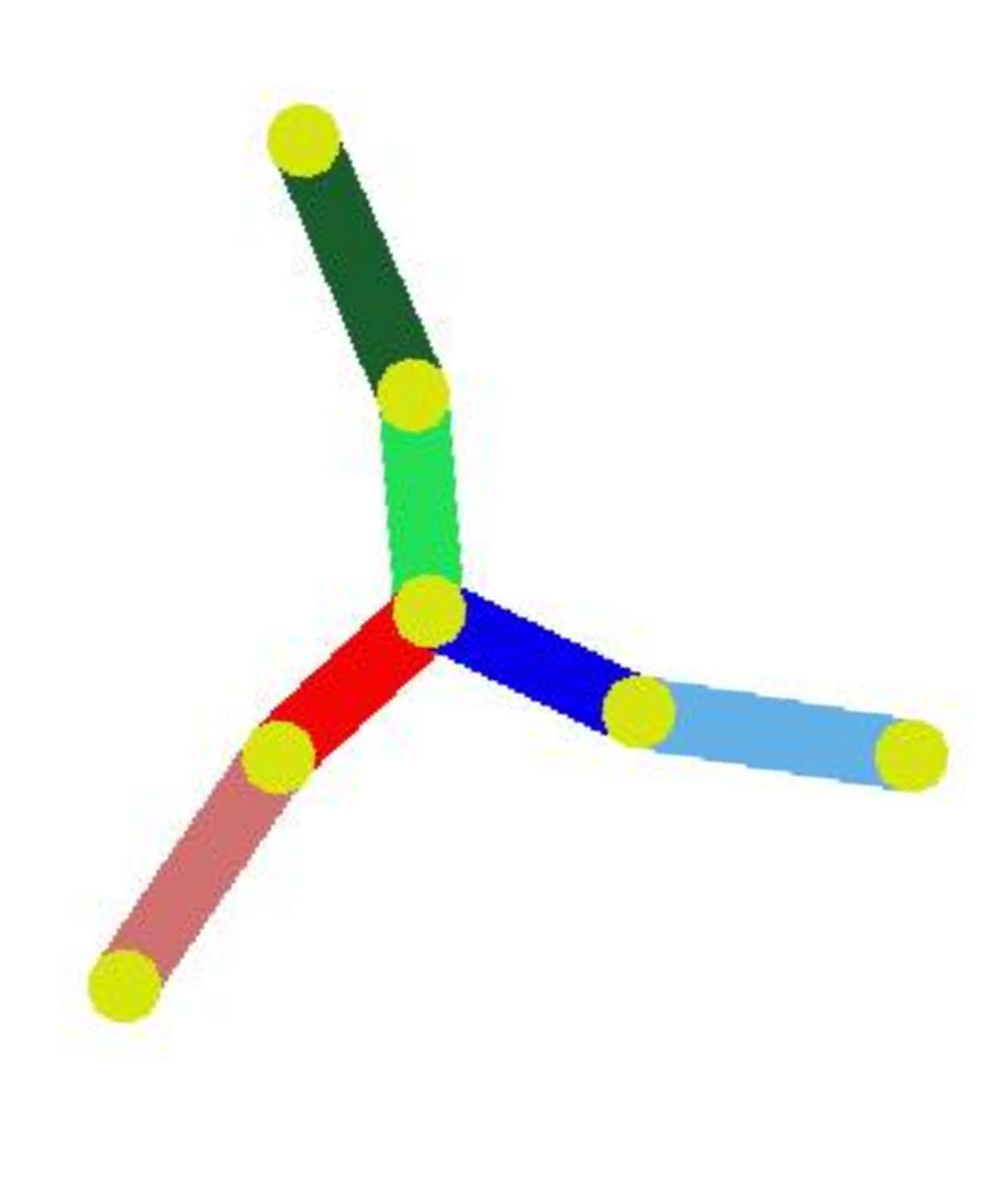}
\caption{}
\label{fig:spider_example}
\end{subfigure}\hfill%
\begin{subfigure}[b]{3.95cm}
\includegraphics[height=3.95cm,keepaspectratio]{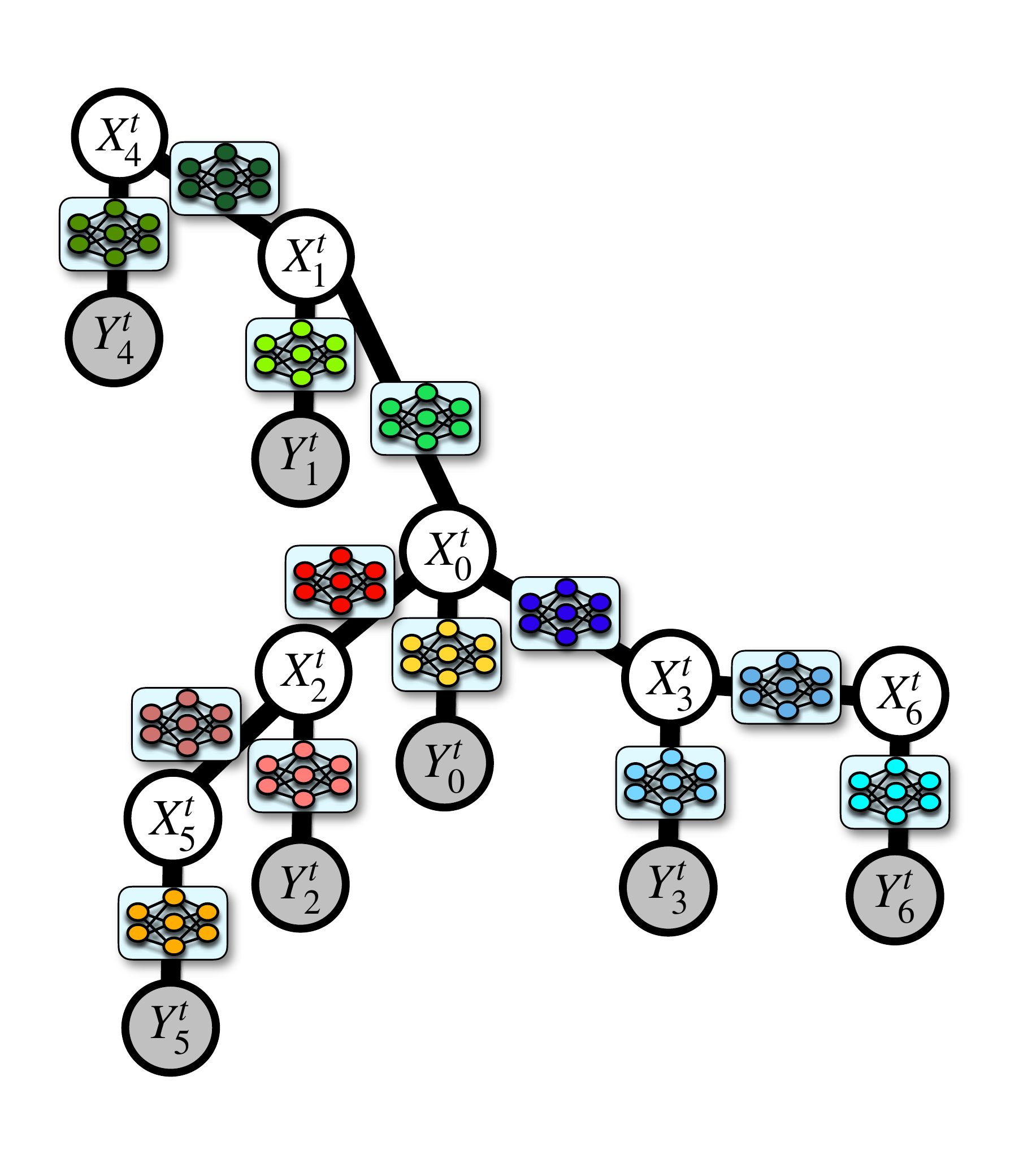}
\caption{}
\label{fig:spider_model}
\end{subfigure}
\caption{\footnotesize a) Geometry and example configuration of the double pendulum. b) Graphical model used by DNBP for the double pendulum task. c) Geometry and an example configuration of the spider structure. d) Graphical model used by DNBP for the spider task.}
\label{fig:toy_models}
\end{center}
\end{figure}

Consider a Markov Random Field (MRF) defined by the undirected graph $\mathcal{G}=\{\mathcal{V}, \mathcal{E}\}$, where $\mathcal{V}$ denotes a set of nodes and $\mathcal{E}$ denotes a set of edges. An example MRF model is shown in \cref{fig:pendulum_model}. Each node in $\mathcal{V}$ represents an observed (grey) or unobserved (white) random variable, while each edge in $\mathcal{E}$ represents a pairwise relationship between two random variables in $\mathcal{V}$. The joint probability distribution for $\mathcal{G}$ is:
\begin{equation}
\label{eq:jointprob}
    p(\mathcal{X}, \mathcal{Y}) = \frac{1}{Z} \prod_{(s,d)\in \mathcal{E}}\psi_{sd}(X_s, X_d) \prod_{d \in \mathcal{V}}\phi_{d}(X_d, Y_d)
\end{equation}
where $\mathcal{X} = \{X_d\,\vert\,d\in \mathcal{V}\}$ is the set of unobserved variables and $\mathcal{Y} = \{Y_d\,\vert\,d\in \mathcal{V}\}$ is the set of corresponding observed variables. The scalar $Z$ is a normalizing constant. 
For each node, the function $\phi_{d}(\cdot)$ is the \textit{unary potential}, describing the compatibility of $X_d$ with a corresponding observed variable $Y_d$.
For each edge, the function $\psi_{sd}(\cdot)$ is the \textit{pairwise potential}, describing the compatibility of neighboring variables $X_s$ and $X_d$. This work considers MRF models limited to pairwise clique potentials. 

Given the factorization of the joint distribution defined in \cref{eq:jointprob}, BP provides an algorithm for inference of the marginal posterior distributions, know as the beliefs, $bel_d(X_d)$.
BP defines a message passing scheme for calculation of the beliefs as follows: 
\begin{equation}
\label{eq:bpbelief}
    bel_d(X_d)\propto \phi_d(X_d,Y_d)\prod_{s\in\rho(d)} m_{s\rightarrow d}(X_d)
\end{equation}
where $\rho(s)$ denotes the set of neighboring nodes of $s$. A message from node $s$ to $d$ is defined as:
\begin{equation}
\label{eq:bpmsg}
    m_{s\rightarrow d}(X_d)=\int_{X_s} \phi_s(X_s,Y_s) \, \psi_{sd}(X_s,X_d) \times \prod_{u\in\rho(s)\setminus d} m_{u\rightarrow s}(X_s) \;dX_s
\end{equation}
Performing inference of random variables in continuous space causes the integral in \cref{eq:bpmsg} to become intractable. This motivates the use of efficient algorithms that approximate the message passing scheme of \cref{eq:bpbelief} and \cref{eq:bpmsg}.

\subsection{Nonparametric Belief Propagation}
\label{sec:pmpnbp}
Nonparametric belief propagation (NBP) \citep{nbp:SudderthIFW03} uses Gaussian mixtures to represent the beliefs and messages for continuous random variables. Later works, including \citet{ihler2009particle} and \citet{bpposeest:DesinghLOJ19}, further improve upon the tractibility of approximate nonparametric inference by representing beliefs and messages with sets of weighted particles. These particle-based NBP methods infer an approximation of the beliefs using an iterative message passing algorithm, in which beliefs and messages are updated at each iteration $t$. In particular, \citet{bpposeest:DesinghLOJ19} avoid the expensive message generation of NBP by approximating \cref{eq:bpmsg} with a ``pull'' strategy. A message, $m^t_{s\rightarrow d}$, outgoing from $s$ to $d$, is generated by first sampling $M$ independent samples from $bel_d^{t-1}(X_d)$ then reweighting and resampling from this set.

\section{Differentiable Nonparametric Belief Propagation}
\label{sec:dnbp}
We propose a differentiable nonparametric belief propagation (DNBP) method. DNBP maintains a representation of the uncertainty in the estimate by efficiently approximating the marginal posterior distributions encoded in an MRF.
Our method avoids the need to define hand-crafted functions for each domain by modeling the potentials needed for the computation of the distributions with neural networks that are trained end-to-end. This hybrid generative-discriminative approach leverages the strengths of both NBP and neural networks.
%

DNBP uses an iterative, differentiable message passing scheme to infer the beliefs over hidden variables in an MRF. DNBP approximates the belief and messages in \cref{eq:bpbelief} and \cref{eq:bpmsg} at iteration $t$ by sets of $N$ and $M$ weighted particles respectively:
\begin{align}
    bel_d^{t}(X_{d}) &= \left\{\left(\mu_d^{(i)}, w_d^{(i)}\right)\right\}_{i=1}^{N}\label{eq:dnbpbel}\\
    m^{t}_{s\rightarrow d} &= \left\{\left(\mu_{sd}^{(i)}, w_{sd}^{(i)}\right)\right\}_{i=1}^{M}\label{eq:dnbpmsg}
\end{align}
%
DNBP relies on a ``pull'' message passing strategy similar to the one presented by \citet{bpposeest:DesinghLOJ19}. In this strategy, each iteration of the algorithm is defined in terms of a message update step and a belief update step. The message update generates a new set of message particles as a reweighted set of samples from the previous iteration's belief. Crucially, the weights associated with these updated message samples result from learned probabilistic factors as opposed to hand-crafted ones. Following a message update, the belief update combines information that is incoming to each node from the newly generated messages. Pseudocode of DNBP's message and belief update schemes is included in \cref{appendix:alg_pseudo}. The following sections describe the networks used to compute the message and belief updates.


\textbf{Unary Potential Functions:}
According to the factorization of the MRF joint distribution in \cref{eq:jointprob}, each unobserved variable $X_d$, for $d\in \mathcal{V}$, is related to a corresponding observed variable $Y_d$ by the unary potential function $\phi_d(X_d, Y_d)$. 
DNBP models each unary function with a feedforward neural network. The unary potential for a particle, $x_d$, given an observed image, $y_d$, is: 
\begin{align}
  \unary{d}{x_d}{y_d} &= l_d\left(x_d\oplus f_d(y_d)\right)
\end{align}
where $f_d$ is a convolutional neural network, $l_d$ is a fully connected neural network, and the symbol $\oplus$ denotes concatenation of feature vectors. 
Details of network architectures are given in \cref{appendix:network_arch}, \cref{tab:netparams}.

\textbf{Pairwise Potential Functions:}
For any pair of hidden variables, $X_s$ and $X_d$, which are connected by an edge in $\mathcal{E}$, a pairwise potential function, $\pairwise{s}{d}{}{}$, represents the probabilistic relationship between the two variables. DNBP models each pairwise potential using a pair of feedforward, fully connected neural networks, $\pairwise{s}{d}{}{}=\{\psi_{sd}^{\rho}(\cdot), \psi_{sd}^{\sim}(\cdot)\}$. The pairwise \textit{density} network, $\psi_{sd}^{\rho}(\cdot)$, evaluates the unnormalized potential for a pair of particles. The pairwise \textit{sampling} network, $\psi_{sd}^{\sim}(\cdot)$, is used to form samples of node $s$ conditioned on node $d$ and vice versa. Details of network architectures are given in the \cref{appendix:network_arch}, \cref{tab:netparams}. The weight computation is detailed in the pseudocode in \cref{appendix:alg_pseudo}.


\textbf{Particle Diffusion:}
DNBP uses a learned particle diffusion model for each hidden variable, modeled as distinct feedforward neural networks, $\tau^{\sim}_{d}(\cdot)$ for $d\in\mathcal{V}$. This diffusion model replaces the Gaussian diffusion models typically used by particle-based inference methods.
At the outset of message generation at iteration $t$, DNBP's belief particles from iteration $t-1$ are resampled then passed through the diffusion model at the beginning of iteration $t$ to form the messages used to update the distributions at iteration $t$. 

\textbf{Particle Resampling:}
The final operation of the belief update algorithm in NBP is a weighted resampling of belief particles.
This resampling operation is non-differentiable~\citep{pfnet:KarkusHL18,dpf:JonschkowskiRB18}.
It follows that the iterative belief update algorithm is non-differentiable due to the resampling step. 
DNBP addresses the non-differentiability of the belief update algorithm by relocating the resampling and diffusion operations to the beginning of the message update algorithm. With this modification, the belief update returns a weighted set of particles approximating the marginal beliefs. The resulting belief density estimate is differentiable up to the beginning of the message update, when particles from the previous iteration were resampled. 
The resulting algorithm is differentiable through one belief update and message passing updates.









\subsection{Supervised Training}
DNBP's training approach is inspired by the work of \citet{dpf:JonschkowskiRB18} with modifications to enable learning the potential functions distinct to DNBP. During training, DNBP uses a set of observation sequences, and a corresponding set of ground truth sequences. 
Using the observation sequences, DNBP estimates belief of each unobserved variable at each sequence step. Then, by maximizing estimated belief at the ground truth label of each unobserved variable, DNBP learns its network parameters by maximum likelihood estimation. 
Further details regarding the implementation of the training procedure are discussed in \cref{appendix:network_arch}.

\textbf{Objective Function:}
Given a set of weighted particles representing the belief of $X_d$ produced by the inference procedure at iteration $t$, the density of the belief can be expressed as a mixture of Gaussians, with a component centered at each particle. The density of a sample $x_d$ can be computed as follows: 
\begin{equation}
\label{eq:density_eval}
    \overline{bel}_d^{t}(x_d) = \sum_{i=1}^N w_d^{(i)}\cdot \mathcal{N}(x_d; \mu_d^{(i)}, \Sigma)
\end{equation}
DNBP defines a loss function one each hidden node $d\in\mathcal{G}$ as:
\begin{equation}
    L^t_d = -\log(\overline{bel}^t_d(x^{t,*}_d))
\label{eq:loss}
\end{equation}
where $x^{t,*}_d$ denotes the ground truth label for node $d$ at sequence step $t$. The loss for each hidden node is computed and optimized separately.
At each sequence step during training, DNBP iterates through the nodes of the graph, updating each node's incoming messages and belief followed by a single optimization step of \cref{eq:loss} using stochastic gradient descent.

\section{Results}
The capability of DNBP is demonstrated on three challenging articulated tracking tasks. The first two tasks involve visually tracking the articulated joints of simulated articulated structures, as illustrated in \cref{fig:toy_models}. To increase the difficulty of these tasks, simulated clutter in the form of static and dynamic geometric shapes are rendered into the image sequences. In the second task, we evaluate DNBP on its ability to track the articulated pose of human hands. In both experiments, DNBP is directly compared to learned baseline approaches that are not NBP.



\subsection{Datasets}
\label{sec:datasets}



\textbf{Simulated Double Pendulum:} To characterize DNBP's tracking performance under chaotic motion, the double pendulum task was chosen as an initial evaluation. The double pendulum structure consists of two revolute joints connected to two rigid-body links in series (see \cref{fig:pendulum_example} for illustration), which are acted on by gravity. 
The pose of the double pendulum is modeled by the 2-dimensional position of its two revolute joints, rendered as yellow circles, and one end effector. The training set on this task consists of $1024$ total sequences with $20$ frames per sequence while the validation set consists of $150$ total sequences with $20$ frames per sequence. Both training and validation sequences are split evenly among three bins of clutter ratio\footnote{In this work, clutter ratio is defined as the ratio of pixels occluded by simulated clutter to the total number of image pixels and is averaged over a full sequence of images.}: none, $0$ to $0.04$ and $0.04$ to $0.1$. Of the training and validation sequences with any amount of clutter, half contain static clutter and the other half contain dynamic clutter. The held-out test set is evenly split among clutter ratio deciles from $0$ to $0.95$, thus contains a shift in distribution from the training set, which was limited to clutter ratios below $0.1$. Each decile contains $50$ sequences with $100$ frames per sequence. For test sequences with any amount of clutter, half contain static clutter and the other half contain dynamic clutter.


\textbf{Simulated Articulated Spider:} 
The spider task was chosen to further characterize DNBP's performance using a structure with added articulations and a larger graphical model. 
As depicted in \cref{fig:spider_example}, the spider is comprised of three revolute-prismatic joints, three purely revolute joints, and six rigid-body links. 
An example of the spider is shown in \cref{fig:spider_example}, in which the joints are rendered as yellow circles and the rigid-body links are rendered as coloured rectangles. Unlike the double pendulum, which contained a stationary base joint, the spider is not tethered to any position and can move freely throughout the image under simulated joint control.
The training, validation and test set for this task follow the same respective distributions of clutter as were used in the double pendulum datasets. The training set consists of 2,048 total sequences and the validation set consists of $300$ sequences. The training and validation sequences are split evenly among five bins of clutter ratio: none, $0$ to $0.04$ and $0.04$ to $0.1$, $0.1$ to $0.2$ and $0.2$ to $0.3$. There are $20$ frames per sequence in each of the spider datasets. Both simulated tasks use images of size $128\times 128$ pixels. Ground truth keypoint locations are represented as continuous valued coordinates scaled to range of $[-1,+1]$. 

\textbf{First-Person Hand Action Benchmark:} The FPHAB dataset~\citep{FirstPersonAction_CVPR2018} consists of RGB-D image sequences taken from the first-person perspective. Thus, the dataset captures the pose and motion of human hands as they perform typical actions. This is a challenging dataset with extreme occlusions where complete observations of all the finger joints are rare. In total, there are $1175$ distinct sequences and $105459$ individual image frames. Each image is labeled with the $3D$ position of $21$ hand joints (illustration of joint relations shown in center column of \cref{fig:teaser}). The best-performing hand pose estimation baseline proposed by \citet{FirstPersonAction_CVPR2018} is used for comparison in the current study. Just like \citet{FirstPersonAction_CVPR2018}, DNBP uses only depth observations. To ensure fair comparisons with the FPHAB baseline, this study follows the $1$:$1$ cross-subject training protocol as described in FPHAB.

\subsection{Implementation Details}
\label{sec:implementation_details}
On all three tasks, Adam \citep{adamopt:KingmaB14} is used for network optimization with a batch size of $6$ and models are trained until convergence of the validation loss. 
The graphs used by DNBP are shown in \cref{fig:pendulum_model,fig:spider_model,fig:teaser}. While DNBP uses tree-structured graphs in these experiments, the inference strategy is compatible with graphs containing cycles since DNBP uses a loopy message passing scheme.
DNBP is trained using $100$ particles per message and tested using $200$ particles per message. During training, one message update is performed at each sequence step, while two message updates are used at test time. The pairwise density, pairwise sampling and diffusion sampling processes of DNBP are defined over the relative translations between neighboring nodes. The maximum weighted particle from each marginal belief set of DNBP is used during evaluation for comparison with the ground truth. 

On both simulated tasks, DNBP is compared to an LSTM recurrent neural network~\citep{lstm:HochreiterS97}. Both models use image inputs that are normalized channel-wise based on training set statistics. The total number of trainable parameters between LSTM and DNBP were chosen to be similar. For hand tracking, the preprocessing protocol of \citet{A2J}, is followed. Notably, preprocessing on the hand tracking task assumes ground truth bounding boxes to ensure fair comparison with the baseline method published by \citet{FirstPersonAction_CVPR2018}. Similarly, the feature extractor used by DNBP in the following experiments was designed to emulate the feature extractor of compared baseline. Details of network parameters and inspection of learned relationships are included in the Appendices~\ref{appendix:network_arch} and ~\ref{appendix:pairwise_inspection}.

\begin{figure}[b]
\captionsetup{width=.475\textwidth}
\centering
\begin{minipage}[t]{.5\textwidth}
  \centering
  \includegraphics[width=.95\linewidth]{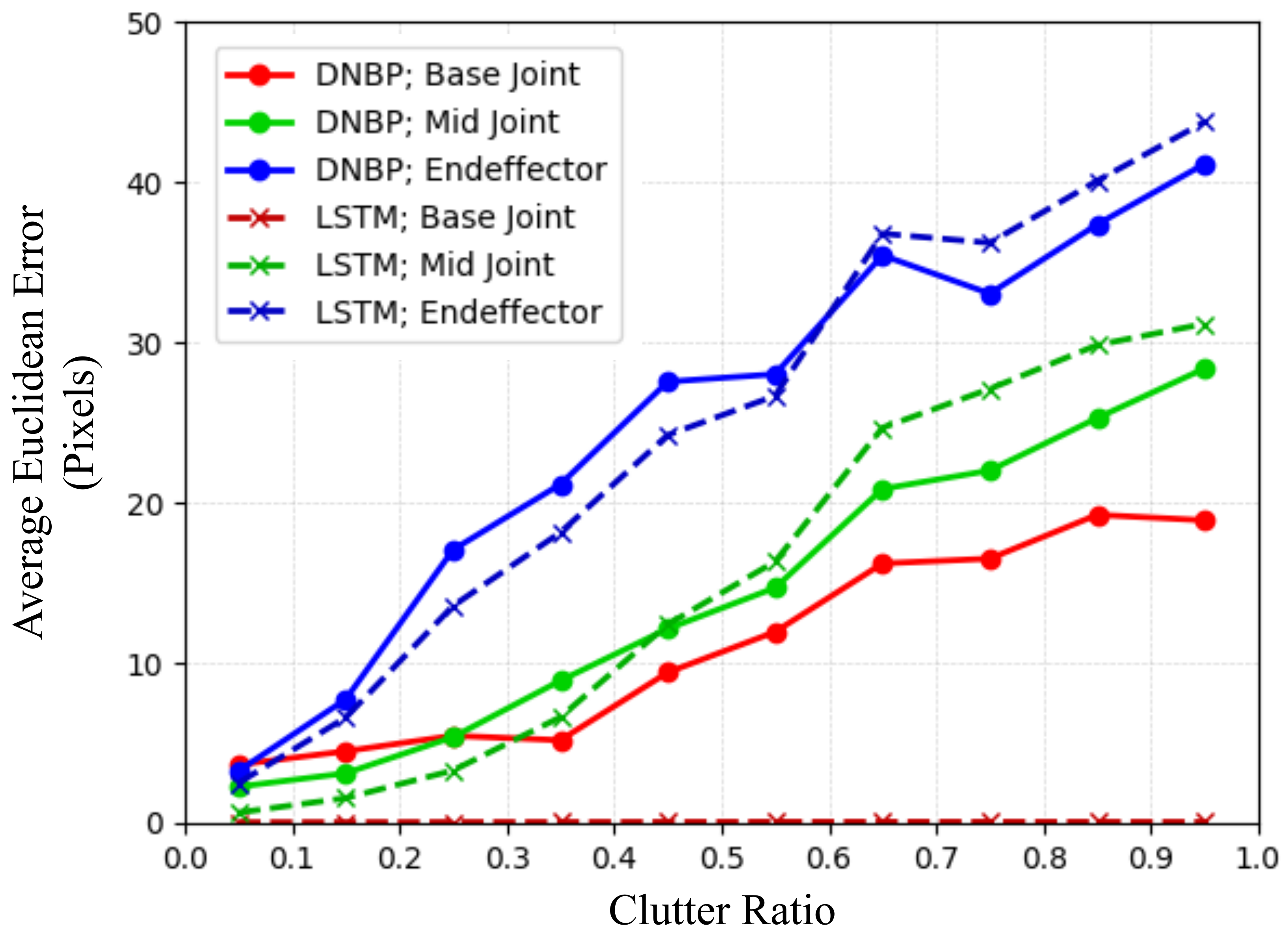}
  \captionof{figure}{Average error of DNBP and LSTM predictions as a function of clutter ratio and keypoint type for double pendulum tracking.}
  \label{fig:pendulum_error}
\end{minipage}%
\begin{minipage}[t]{.5\textwidth}
  \centering
  \includegraphics[width=.95\linewidth]{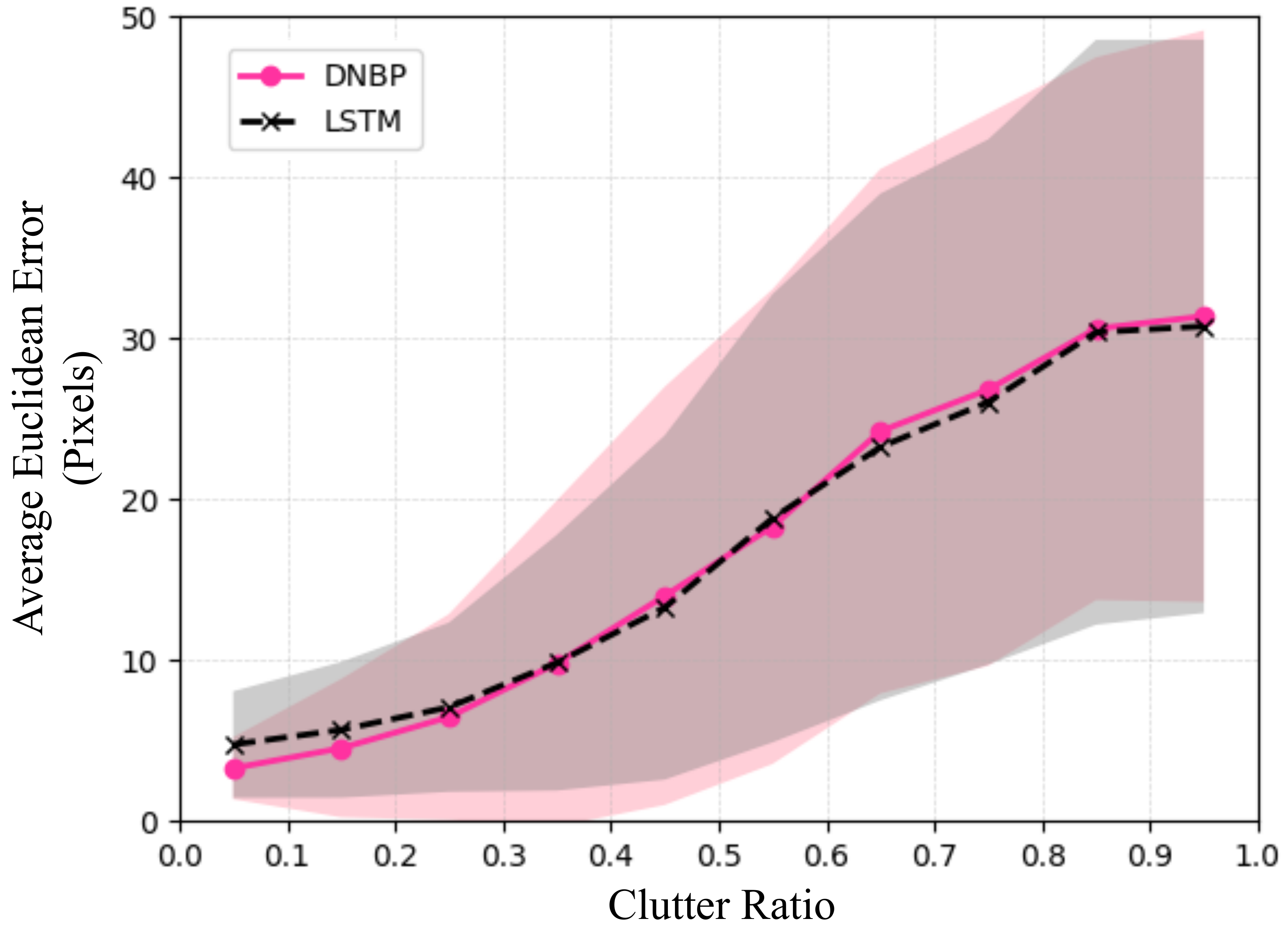}
  \captionof{figure}{Average error of DNBP and LSTM predictions as a function of clutter ratio for articulated `spider' tracking.
  }
  \label{fig:spider_error}
\end{minipage}
\end{figure}

\subsection{Performance metrics}
As a quantitative measure of tracking error, average Euclidean error is used. On the simulated tasks, Euclidean error is averaged over all images in the test set. On the hand tracking task, Euclidean error is averaged over all joints per frame then used to calculate the percent of frames satisfying variable error thresholds as used by~\citet{FirstPersonAction_CVPR2018}.

Discrete entropy \citep{entropy:Shannon48} is used as a quantitative measure of uncertainty estimated by DNBP. Discrete entropy is calculated by binning samples from each marginal belief set.
For qualitative analysis of the uncertainty estimated by DNBP, samples from an approximation of the joint posterior distribution (i.e. for collection of all unobserved variables) are formed using a sequential Monte Carlo sampling approach~\citep{smc:NaessethLS14}. Visualization of these samples are formed by plotting a rendered link between each pair of keypoint samples. 

\subsection{Double Pendulum Tracking Results}
\label{sec:double_pen_results}


As shown in \cref{fig:pendulum_error}, the keypoint tracking error of DNBP is directly compared to that of the LSTM baseline on the held-out test set for each keypoint type (base, middle and end effector) across the full range of clutter ratios. Results from this comparison show that DNBP's average keypoint tracking error is comparable to the LSTM's corresponding error for both the mid joint and end effector keypoints, independent of clutter ratio. For the base joint keypoint, which is stationary at the center position of every image, the LSTM was able to memorize the correct position. DNBP, which diffuses particles based on the message passing scheme, does not memorize the base joint position and registers a consistently larger error which increased with clutter ratio.

\begin{figure}[t]
\captionsetup{width=.475\textwidth}
\centering
\begin{minipage}[t]{.5\textwidth}
  \centering
  \includegraphics[width=.95\linewidth]{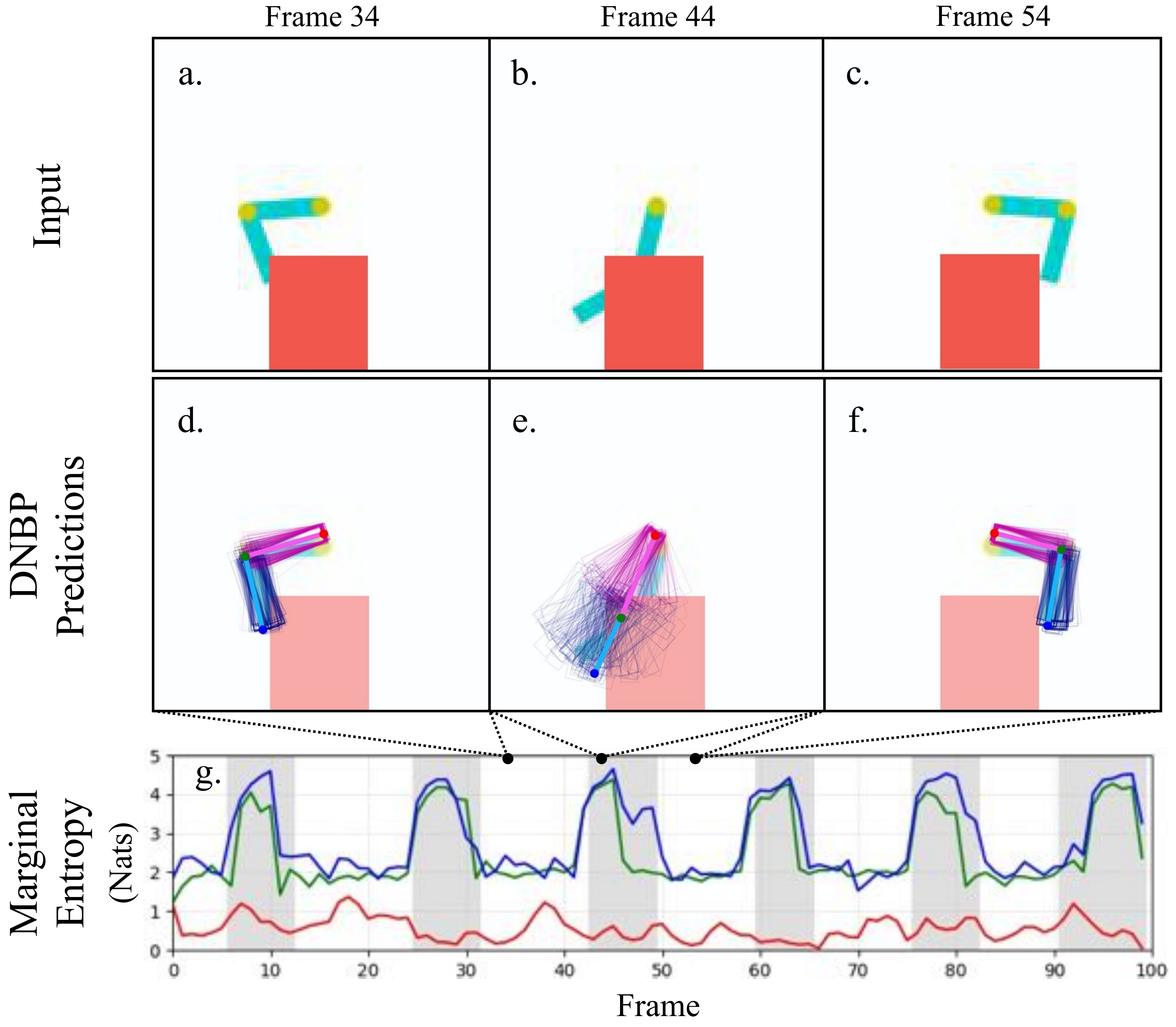}
  \caption{Tracking of double pendulum by DNBP under partial occlusion (orange block). 
  Uncertainty associated with predictions is shown as samples from the joint distribution in pink and blue (d,e,f). 
  (g) Marginal entropy for each keypoint across test sequence; base keypoint (red), middle keypoint (green), end-effector keypoint (blue). Sequence steps highlighted by gray correspond to images in which $>25\%$ of the pendulum is occluded.}
  \label{fig:pendulum_occ}
\end{minipage}%
\begin{minipage}[t]{.5\textwidth}
  \centering
  \includegraphics[width=0.95\linewidth]{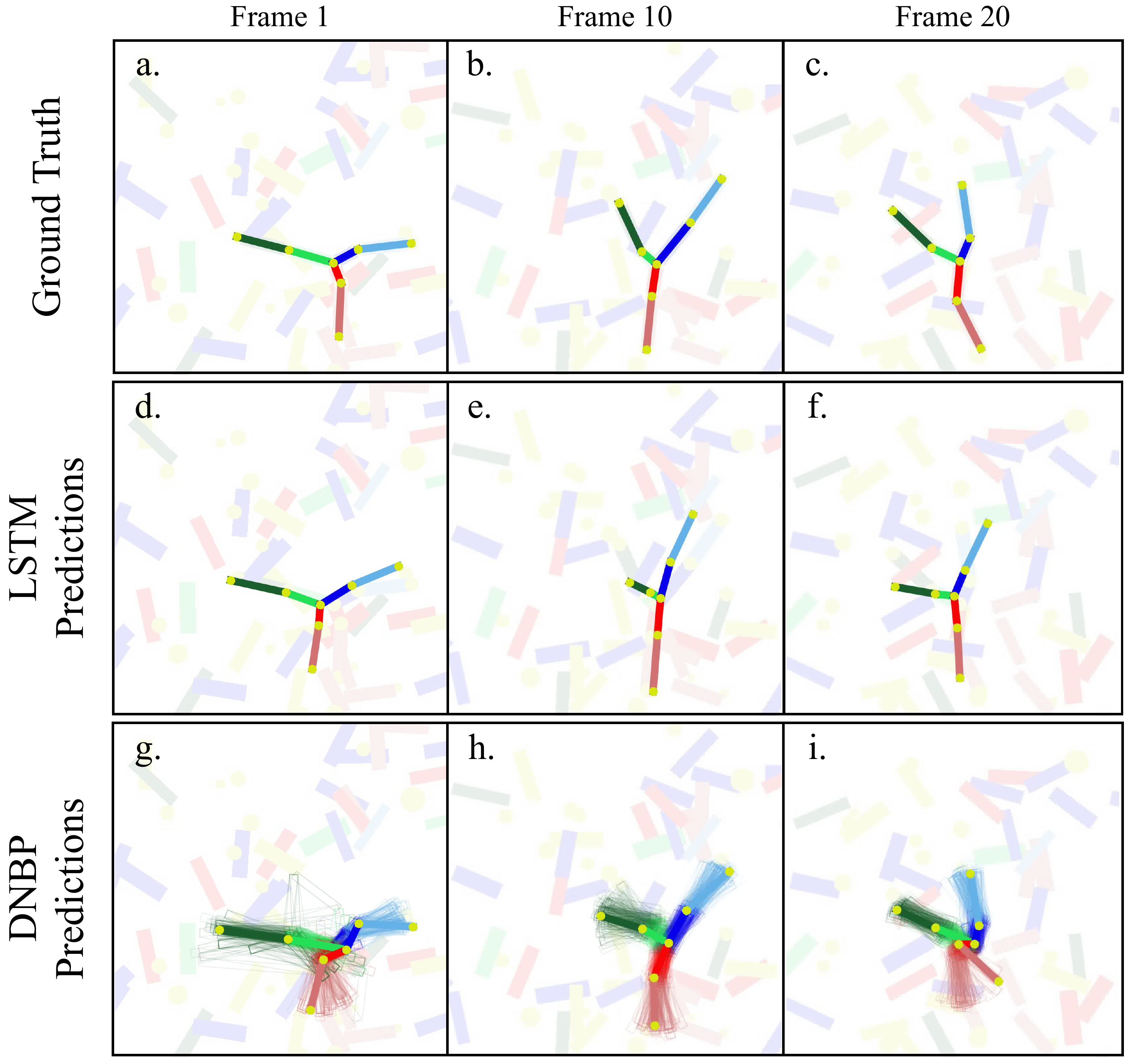}
  \caption{Comparison of articulated `spider' tracking by LSTM (d,e,f) and DNBP (g,h,i) under cluttered conditions. 
Predicted and ground truth keypoints shown as yellow circles. Clutter shown as faded shapes for illustration to highlight predictions.}
  \label{fig:spidertrack}
\end{minipage}
\end{figure}

DNBP provides measures of uncertainty associated with its predictions, which are generated according to the algorithmic prior of belief propagation. Next tested was the hypothesis that the DNBP model would generate increased uncertainty under conditions in which an occluding object is placed into the input images such that it covers portions of the double pendulum. This test was performed by rendering an occluding block onto a test sequence as shown in \cref{fig:pendulum_occ}a-c. Under optimal conditions, in which the pendulum is minimally occluded ($<25\%$ by surface area), the model's output indicates a low level of uncertainty (see \cref{fig:pendulum_occ}d,f,g.) for each keypoint and each frame. In contrast, under conditions in which the pendulum is occluded by the superimposed object, the model's output indicates relatively high levels of uncertainty precisely at frames in which the superimposed object occludes a portion ($>25\%$) of the double pendulum (see \cref{fig:pendulum_occ}e,g.). These results demonstrate that the estimate of uncertainty produced by DNBP can identify predictions which are unreliable.


\subsection{Articulated Spider Tracking Results}
\label{sec:spider_results}

After having established the performance characteristics of DNBP on the relatively straightforward double pendulum task, we next set out to determine DNBP's capability for tracking more complex structures. To this end, the 3-arm spider structure is used as a more challenging articulated pose tracking task. Each model's performance is quantitatively assessed on the held-out test set of the articulated spider tracking task using the same approach as described for the double pendulum experiment by varying clutter ratio (\cref{fig:spider_error}). Similar to the results of the double pendulum experiment, average error on the spider task increases as a function of clutter ratio for both the LSTM and for DNBP. For clutter ratios between $0$ and $0.25$, average error for both models remains near $6$ pixels then increases consistently with clutter ratio, reaching above $30$ pixels of average error for clutter ratios above $0.85$. As in the case of the double pendulum experiment, these results demonstrate comparable performance between LSTM and DNBP on an articulated pose tracking task.


Next, a qualitative example of tracking performance under conditions of clutter is shown in \cref{fig:spidertrack}. In \cref{fig:spidertrack}(a-c), the ground truth pose is shown amidst distracting shapes across selected frames of a test sequence with clutter ratio of $0.25$. Pose predictions generated by LSTM are shown in \cref{fig:spidertrack}(d-f) and by DNBP in (g-i). Qualitative assessment of the images indicates both the LSTM and DNBP place their predictions in the correct region of the image. Additionally, each model is shown to correctly predict the relative positions of the three arms. Over the sequence, both models track the motion of each keypoint, however appear to struggle with certain keypoint predictions.

\begin{figure}[t]
\centering
\includegraphics[width=0.89\linewidth]{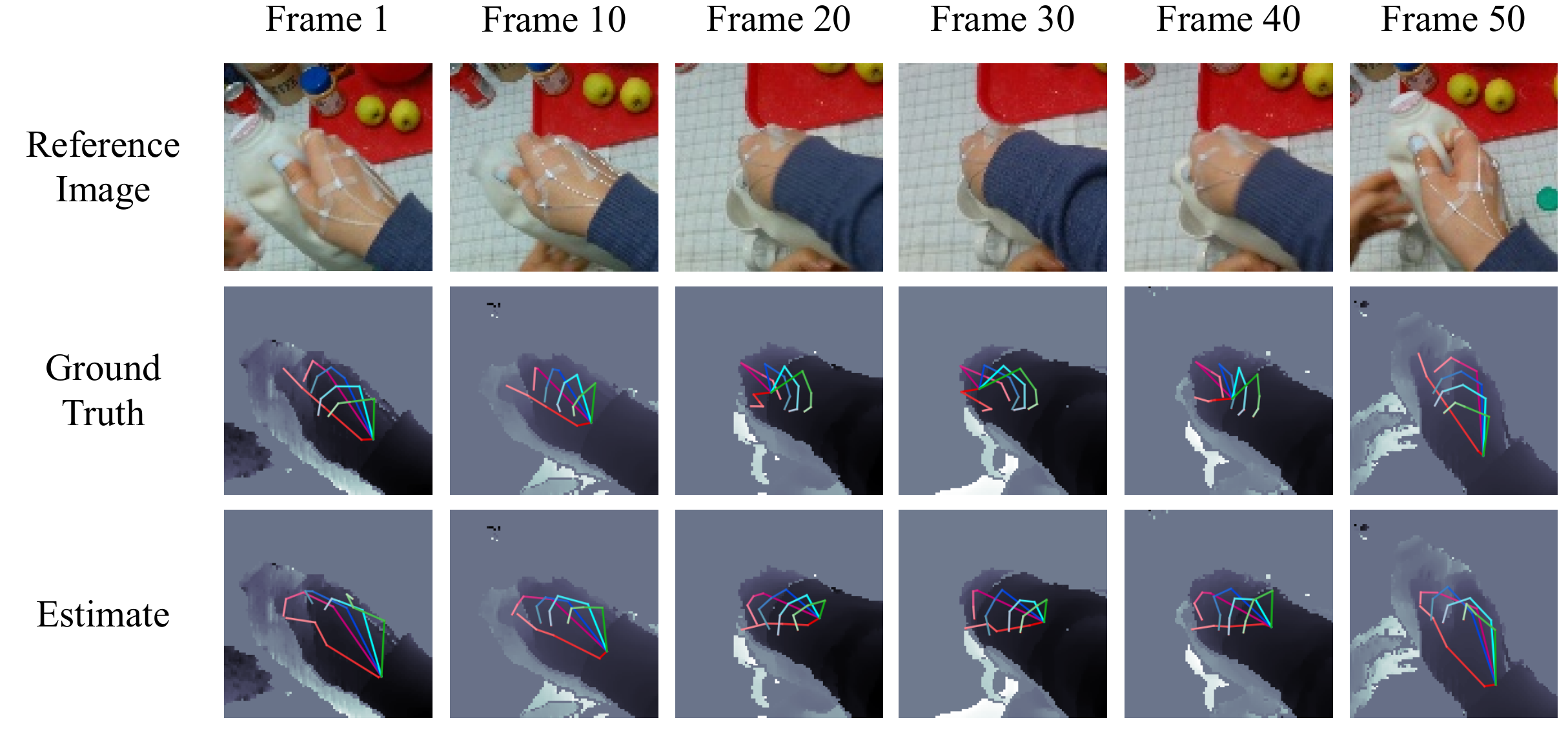}
\caption{\footnotesize Output from DNBP throughout a chosen sequence of hand tracking. DNBP maintains plausible estimates of the hand pose in cases of occlusion (Frames 20, 30, 40) and recovers with improved observability (Frame 50).}
\label{fig:hand_qual_seq}
\end{figure}

\begin{figure}[t]
\centering
\includegraphics[width=0.89\linewidth]{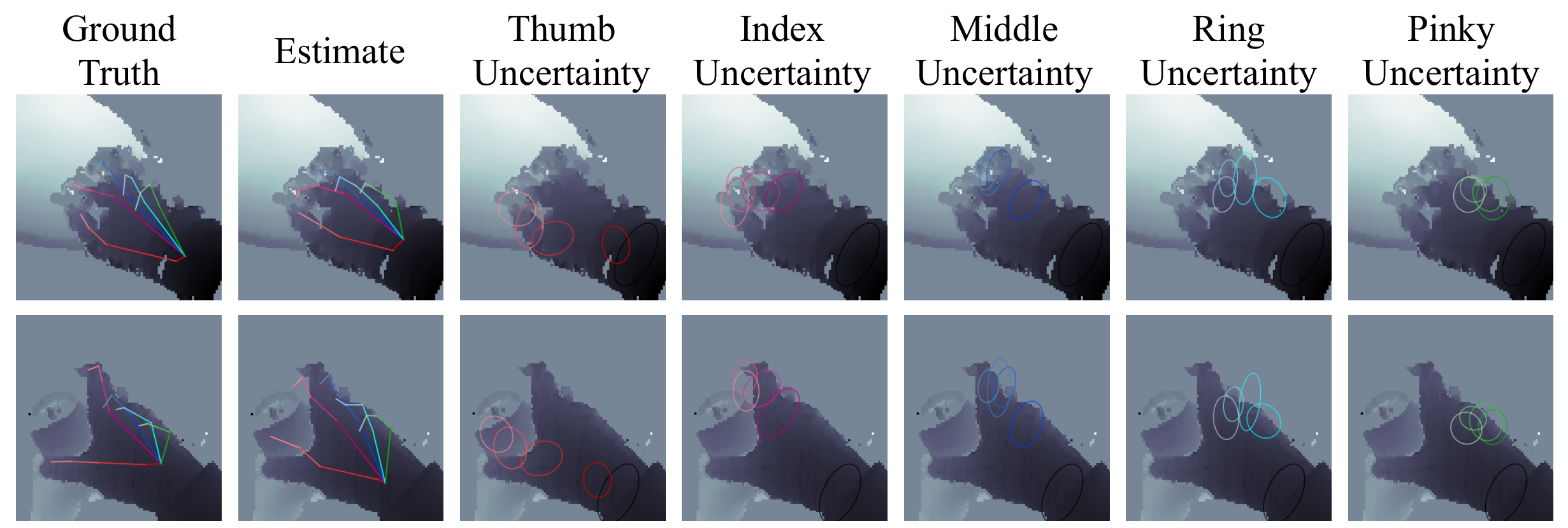}
\caption{\footnotesize Output from DNBP on randomly sampled frames. See \cref{appendix:hand_tracking_resutls} for more examples.}
\label{fig:hand_qual}
\end{figure}

\subsection{Human Hand Tracking Results}
\begin{figure}[b]
\centering
\includegraphics[width=7cm]{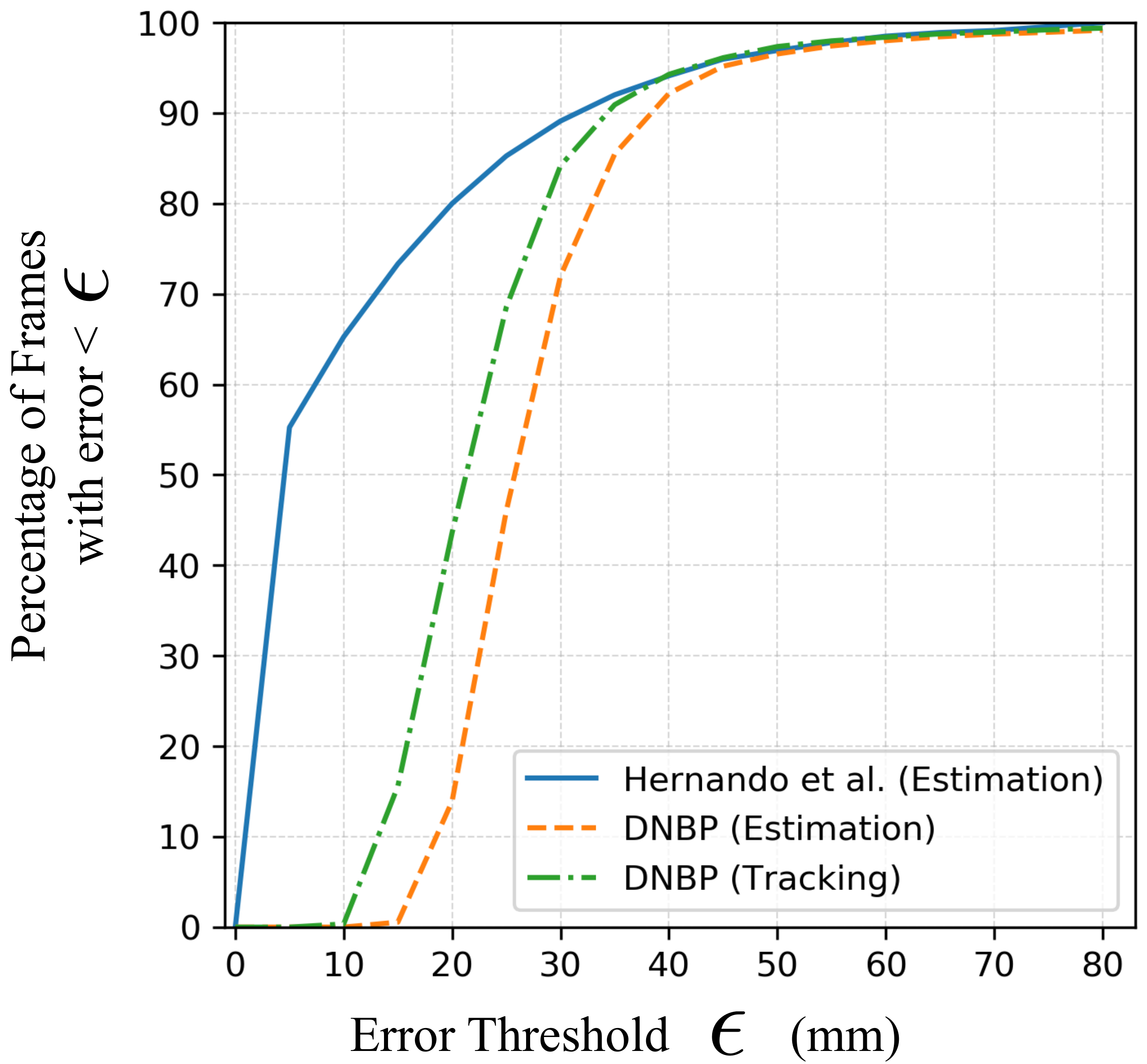}
\caption{\footnotesize Quantitative comparison between DNBP and neural network baseline on hand pose tracking task of the FPHAB dataset. For each model the percent of frames with predicted pose less than a set threshold is calculated as the threshold is varied from $0$mm to $80$mm.}
\label{fig:hand_error}
\end{figure}

To evaluate DNBP's capability for application to real-world tasks, the algorithm's state estimation and tracking performance was evaluated on the FPHAB dataset. This is a challenging dataset with extreme occlusions where complete observations of all the finger joints are rare. Firstly, Euclidean error between the estimated and ground truth pose is measured for every frame in the test set. For this first evaluation, DNBP is applied as a frame-by-frame estimator without maintaining its belief over time. The quantitative results from this experiment, are included in \cref{fig:hand_error} with direct comparison to a pure neural network baseline. The results from this experiment indicate that for error thresholds below $50$mm, DNBP will consistently have an accuracy of 95\% and above. 

Following the comparison against a state of the art baseline, it was hypothesized that DNBP's performance would improve when applied as a tracking method which maintains belief over time. To perform this test, DNBP was applied sequentially to each test sequence and evaluated under the same error metric. 
The result from this test, as shown in \cref{fig:hand_error}, demonstrates that DNBP does improve in terms of frame error when allowed to track its uncertainty over time. 
Qualitative examples (on frames from randomly chosen sequences) showing DNBP's tracking performance are shown in~\cref{fig:hand_qual_seq,fig:hand_qual} and Appendix~\ref{appendix:hand_tracking_resutls}. 
The tracking videos showing the DNBP's estimates and belief are included in the supplementary material and project webpage: \href{https://progress.eecs.umich.edu/projects/dnbp/}{https://progress.eecs.umich.edu/projects/dnbp/}.


\section{Conclusion}
\label{sec:discussion}

In this work, we proposed a novel formulation of belief propagation which is differentiable and uses a nonparametric representation of belief. It was hypothesized that combining maximum likelihood estimation with the nonparametric inference approach would enable end-to-end learning of the probabilistic factors needed for inference. Results on both qualitative and quantitative experiments demonstrate successful application of this approach and highlight the capability of DNBP to estimate useful measures of uncertainty, which are crucial for applications where incorrect estimates lead to catastrophic decisions, such as robotics.
The current approach is limited by its use of non-differentiable resampling and its demand for a graph model as input. 
Exploration of methods to overcome these limitations, such as by incorporating a soft-resampling strategy~\cite{pfnet:KarkusHL18}, are left as future work.
Inspired by recent work that has extended differentiable state estimation algorithms into the planning domain~\citep{dancontrol:KarkusMHKLL19,dualsmc:WangL0ZD0T20,ghostplan:AndersonSPBL19}, we see the potential to embed DNBP within a differentiable planning system as an exciting direction for future work.

\bibliographystyle{ACM-Reference-Format}
\bibliography{dnbpbib}

\clearpage
\section{Appendix}
\label{appendix}
\subsection{Algorithm Pseudocode}
\label{appendix:alg_pseudo}
In this section, pseudo code of the DNBP message passing algorithm is given for reference. As discussed in \cref{sec:dnbp}, this algorithm is a differentiable variant of the PMPNBP algorithm~\cite{bpposeest:DesinghLOJ19}. 

Results in this work were generated with $U$ set to $10$, while past related work~\citep{bpposeest:DesinghLOJ19} used $U=1$. It was observed that this modification improved training stability during preliminary development. Note that $\gamma$ is a hyperparameter that controls the resampling strategy and is set to $0.9$ in our experiments. $\gamma$ is used only during training; during evaluation, all $M$ samples are drawn from $\bel{d}{n-1}$.


\begin{algorithm}[h]
\SetAlgoLined
\SetKwInOut{Input}{input}\SetKwInOut{Output}{output}
\Input{Belief set $\bel{d}{n-1}=\set{(\wgt{d}{}{i},\prt{d}{}{i})}_{i=1}^{T}$\newline Incoming messages $\msg{u}{s}{n-1}=\set{(\wgt{u}{s}{i},\prt{u}{s}{i})}_{i=1}^{M}$ for each node $u\in \rho(s)\setminus d$}

\Output{Outgoing messages, $\msg{s}{d}{n}=\set{(\prt{s}{d}{i},\wgt{s}{d}{i})}_{i=1}^{M}$}
\BlankLine
 Draw $(1-\gamma^{n-1})\cdot M$ independent samples from $\bel{d}{n-1}$\newline$\set{\prt{s}{d}{i}\gets \bel{d}{n-1}}_{i=1}^{(1-\gamma^{n-1})\cdot M}$\;
 Apply particle diffusion to each sampled particle\newline$\prt{s}{d}{i}=\prt{s}{d}{i}+\temporal{d}$\;
 Draw remaining $\gamma^{n-1}\cdot M$ samples independently from uniform proposal distribution\;
 
 \ForEach{$\set{\prt{s}{d}{i}}_{i=1}^{M}$}{
    \For{$\ell=[1:U]$}{
    Sample $\hat{X_s}^{(i)}\sim \pairwise{s}{d}{\prt{s}{d}{i}}{}$\;
    $w_{unary}^{(i)} = w_{unary}^{(i)} + \unary{s}{\hat{X_s}^{(i)}}{}$\;
    }
    
    $w_{unary}^{(i)} = \frac{w_{unary}^{(i)}}{U}$\;
    
    \ForEach{$u\in \rho(s)\setminus d$}{
        $W_u^{(i)}=\sum\limits_{j=1}^{M}\wgt{u}{s}{j}\times w_{u}^{(ij)}$ where $w_{u}^{(ij)}=\pairwise{s}{d}{\prt{u}{s}{j}}{\prt{s}{d}{i}}$\;
    }
    $w_{neigh}^{(i)} = \prod\limits_{u\in \rho(s)\setminus d} W_u^{(i)}$\;
    $\wgt{s}{d}{i}=w_{unary}^{(i)}\times w_{neigh}^{(i)}$\;
 }
 
 Associate $\set{\wgt{s}{d}{i}}_{i=1}^{M}$ with $\set{\prt{s}{d}{i}}_{i=1}^{M}$ to form outgoing $\msg{s}{d}{n}$\;
 \caption{\label{alg:msgupdate} Message update}
\end{algorithm}

\clearpage

\begin{algorithm}[t]
\SetAlgoLined
\SetKwInOut{Input}{input}\SetKwInOut{Output}{output}
\Input{Incoming messages, $\msg{s}{d}{n}=\set{(\wgt{s}{d}{i},\prt{s}{d}{i})}_{i=1}^{M}$, for each node $s\in \rho(d)$}
\Output{Belief set $\bel{d}{n}=\set{(\wgt{d}{}{i},\prt{d}{}{i})}_{i=1}^{T}$}
\BlankLine
\ForEach{$s\in \rho(d)$}{
 Update message weights \newline$\wgt{s}{d}{i} = \wgt{s}{d}{i} \times \unary{d}{\prt{s}{d}{i}}{}$ for $i\in[1:M]$\;

 Normalize message weights \newline$\wgt{s}{d}{i} = \frac{\wgt{s}{d}{i}}{\sum\limits_{j=1}^{M} \wgt{s}{d}{j}}$ for $i\in[1:M]$\;
 }
 Form belief set $\bel{d}{n}=\bigcup\limits_{s\in \rho(d)} \msg{s}{d}{n}$\;
 Normalize belief weights \newline$\wgt{d}{}{i} = \frac{\wgt{d}{}{i}}{\sum_{j=1}^{T} \wgt{d}{}{j}}$ for $i\in[1:T]$\;
 
 \caption{\label{alg:beliefupdate} Belief update}
\end{algorithm}

\subsection{Network Architecture \& Training}
\label{appendix:network_arch}
For both simulated articulated tracking tasks, the network architecture for each sub-network described in \cref{sec:dnbp} is summarized in \cref{tab:netparams}. For the hand tracking task, each network follows the same structure as those in \cref{tab:netparams}, with two exceptions: (1) the feature extractor, $f_s(\cdot)$, used for hand tracking is based on the architecture used by the FPHAB baseline that was introduced by~
\citet{Ye2016SpatialAD}. (2) each node likelihood network, $l_s(\cdot)$, has one additional feature reduction layer of [fc(64, BatchNorm,ReLU)] preceeding the layers of the corresponding network in \cref{tab:netparams}.

The following sections describe specific implementation details used in the supervised training of DNBP. To ensure independence from spatial location, the pairwise density, pairwise sampling and diffusion sampling processes of DNBP are defined over the space of transformations between variables. Specifically, each of these networks takes as input or produces as output a translation between samples of their corresponding random variables.

\begin{table}[b]
\centering
\begin{tabular}{ c l }
\textbf{NETWORK}  & \multicolumn{1}{c}{\textbf{UNIT LAYERS}}\\
\hline \\

$f_s$    & 5 x [conv(3x3, 10, stride=2, ReLU), maxpool(2x2, 2)]\\ 

$l_s$    & 2 x fc(64, ReLU), fc(1, Sigmoid scaled to [0.005, 1])\\ 

$\psi_{sd}^{\rho}$    & 4 x fc(32, ReLU), fc(1, Sigmoid scaled to [0.005, 1])\\ 

$\psi_{sd}^{\sim}$    & 2 x fc(64, ReLU), fc(2)\\ 

$\tau^{\sim}_{s}$    & 2 x fc(64, ReLU), fc(2)\\ 
\end{tabular}
\caption{Network parameters of learned DNBP potential functions used on both simulated articulated tracking tasks. Note $s,d\in \mathcal{V}$, and $(s,d)\in \mathcal{E}$. Unary potentials: $l_s(f_s(\cdot))$. Pairwise potentials: $\{\psi_{sd}^{\rho}, \psi_{sd}^{\sim}\}$. Particle diffusion: $\tau^{\sim}_{s}$.}
\label{tab:netparams}
\end{table}

\textbf{Gradient Decoupling:}
The belief weight, $w_d^{t,(i)}$, of particle $i$ is proportional to the product of \textit{component} weights, $w^{t,(i)}_{unary_d}\times w^{t,(i)}_{unary_s}\times w^{t,(i)}_{neigh_s}$, where $s$ is the neighbor of node $d$ from which particle $i$ originated (see Algorithm \ref{alg:beliefupdate}). Since each of these component weights is produced by a separate potential network (either $\phi_d$, $\psi_{sd}^{\sim}$, or $\psi_{sd}^{\rho}$ respectively), direct optimization of the belief density will lead to interdependence of the potential network gradients during training. In the context of DNBP, interdependence between different potential functions is inconsistent with the factorization given in $\mathcal{G}$. \citet{cnngraph:TompsonJLB14} describe a similar phenomenon they refer to as gradient coupling which was addressed by expressing a product of features in log-space which ``decouples'' the gradients.

To avoid interdependence between potential functions during training, we consider the \textit{partial}-belief densities which are defined for each node $d\in \mathcal{V}$ as mixtures of Gaussian density functions:
\begin{align}
    &\overline{bel}^{t}_{d,unary_d}(X_d) = \sum_{i=1}^N w^{t,(i)}_{unary_d}\cdot \mathcal{N}(X_d; \mu_d^{(i)}, \Sigma) \label{eq:pseudobellik} \\
    &\overline{bel}^{t}_{d,unary_{\rho(d)}}(X_d) = \sum_{i=1}^N w^{t,(i)}_{unary_s}\cdot \mathcal{N}(X_d; \mu_d^{(i)}, \Sigma) \label{eq:pseudobelunary} \\
    &\overline{bel}^{t}_{d,neigh_{\rho(d)}}(X_d) = \sum_{i=1}^N w^{t,(i)}_{neigh_s}\cdot \mathcal{N}(X_d; \mu_d^{(i)}, \Sigma) \label{eq:pseudobelneigh}
\end{align}
Using these definitions, direct interaction between the potential networks' gradients is avoided by maximizing the product of partial-beliefs at the ground truth of each node in log space. The product of partial-beliefs is defined:
\begin{equation}
    \overline{bel}^t_d(X_d) = \;\overline{bel}^{t}_{d,unary_d}(X_d)\times \overline{bel}^{t}_{d,unary_{\rho(d)}}(X_d) \times \overline{bel}^{t}_{d,neigh_{\rho(d)}}(X_d)
\label{eq:pseudbel}
\end{equation} 

\textbf{Unary Potentials:}
During training of DNBP, only those gradients derived from the belief update of each node are used to update the corresponding node's unary potential network parameters. Any gradients derived from the outgoing messages of a particular node are manually stopped from propagating to that node's unary network. This is done to avoid confounding the objective functions of neighboring nodes, which each rely on the others' unary network during message passing. This approach can be implemented with standard deep learning frameworks by dynamically stopping the parameter update of each unary network depending on where in the algorithm its forward pass was registered.

\textbf{Pairwise Density Networks:}
To speed up and stabilize the training of pairwise density potential networks, the following substitution is made during training. While calculating $w_{sd}^{(i)}$ for outgoing message $i$ from node $s$ to $d$, the summation over incoming messages from $u\in \rho(s)$ to $s$ is replaced by a single evaluation of:
\begin{equation}
W_u^{(i)}=\pairwise{s}{d}{x^{*}_s}{\prt{s}{d}{i}}
\end{equation}
where $x^{*}_s$ is the ground truth label of sender node $s$. This change improves inference time and reduces memory demands by removing a summation over $M$ particles while also providing more stable training feedback to the network. This substitution is removed at test time after training is complete.

\textbf{Pairwise Sampling Networks:}
The pairwise sampling networks, $\psi_{s,d}^{\sim}$, take a random sample of Gaussian noise as input and generate conditional samples using the following rule:
\begin{align}
    \epsilon&\sim\mathcal{N}(0,1)\\
    x_{s\vert d}&= x_d + \psi_{s,d}^{\sim}(\epsilon)
\end{align}
where $x_{s\vert d}$ is the sample of variable $X_s$ conditioned on neighboring sample $x_d$ and where $\epsilon$ is a noise vector sampled from a zero-mean, unit variance multivariate Gaussian distribution with $dim(\epsilon)=64$. Similarly, for sampling in the opposite conditioning direction (node $d$ conditioned on $s$), memory efficiency is gained by reusing the $\psi_{s,d}^{\sim}$ network but negating the sampled translation.

\subsection{Double Pendulum Clutter}
\label{appendix:pendulum_clutter}
As summarized in \cref{sec:datasets}, the double pendulum dataset was generated using a modified version of the OpenAI~\cite{openai:1606.01540} Acrobot environment. Synthetic geometric shapes are rendered into each image of the dataset to simulate noisy, cluttered environments. All simulated clutter on the double pendulum task is generated according to the following parameters: $50\%$ of clutter is rendered visually beneath the pendulum while the remaining $50\%$ is rendered on top of the pendulum. For dynamic clutter, each geometry simulates motion using a random, constant position velocity ($\dot{x}$, $\dot{y}$) and orientation velocity ($\dot{\theta}$). Position velocities are sampled from $\mathcal{N}(0, 0.025)$. Orientation velocities are sampled from $\mathcal{N}(0, 0.05)$. Clutter is simulated as either rectangles with $80\%$ probability or circles with $20\%$ probability. Clutter rectangles are sized randomly with length of $\text{max}(0, l\sim\mathcal{N}(0.2, 0.05))$ and height of $\text{max}(0, h\sim\mathcal{N}(0.8, 0.2))$. Color of clutter rectangles is randomly chosen with RGB of $(0,204,204)$ or $(245,87,77)$. Clutter circles are sized randomly with radius of $\text{max}(0, r\sim\mathcal{N}(0.1, 0.1))$ and colored randomly with RGB of $(204,204,0)$ or $(96,217,63)$. Size and color distributions were chosen to ensure clutter visually resembles the double pendulum parts. The position of each clutter geometry was randomly initialized within $1.5$x the extent of the image boundary.

The training and validation datasets were distributed evenly among clutter ratios of [$0$, $0-0.04$, and $0.04-0.1$]. For the training/validation sequences that included clutter, the number of clutters rendered beneath and on top of the double pendulum was individually randomly sampled from independent Binomial distributions using $n=15$, $p=0.3$. To generate the test set, which was uniformly distributed among clutter ratios as described in \cref{sec:datasets}, rejection sampling was used with variable numbers of rendered geometries.

\subsection{Articulated Spider Model}
\label{appendix:spider_model}
Data for the articulated spider tracking task of \cref{sec:spider_results} was simulated using the Pillow~\cite{clark2015pillow} image processing library. Three revolute-prismatic joints are all centrally located and treated as the root of the spider's kinematic tree. The remaining three revolute joints are attached to pairs of links, forming three distinct 'arms' of the spider. Each joint is rendered as a yellow circle while the six rigid-body links are rendered as distinct red, green or blue rectangles respectively. Size parameters that follow are with respect to rendered image size of  $500$x$500$px. The three inner revolute-prismatic joints include rotational constraints limiting each to a non-overlapping $120^{\circ}$ range of articulation as well as prismatic constraints limiting the extension to within $[20,80]$ pixels of translation. The three purely revolute joints are constrained to rotations between $\pm35^{\circ}$ with respect to their local origins. Each rigid-body link has width of $20$px and height of $80$px pixels while each joint has radius of $10$px. 

For every simulated sequence, the spider is initialized with uniformly random root position within the central $180$x$180$px window and uniformly random root orientation from [$0,2\pi$]. Furthermore, each joint state is initialized uniformly at random within its particular articulation constraints. The spider is simulated with dynamics using randomized, constant root, and joint velocities with respect to a time step ($dt$) of $0.01$. The root's position velocities ($\dot{x}$, $\dot{y}$) are each sampled from an equally weighted $2$-component Gaussian mixture with means ($+24$, $-24$) and standard deviations ($15$, $15$). Whereas, the root's orientation velocity ($\dot{\theta}$) is sampled from an equally weighted $2$-component Gaussian mixture with means ($+0.3$, $-0.3$) and standard deviations ($0.1$, $0.1$). Each rotational joint's velocity is sampled from an equally weighted $2$-component Gaussian mixture with means ($+0.3$, $-0.3$) and standard deviations ($0.1$, $0.1$). Similarly, each prismatic joint's velocity is sampled from an equally weighted $2$-component Gaussian mixture with means ($+500$, $-500$) and standard deviations ($60$, $60$). Note that if any joint reaches an articulation limit during simulation, the direction of its velocity is reversed.

\subsection{Articulated Spider Clutter}
\label{appendix:spider_clutter}
Clutter generation for the articulated spider tracking task follows a similar generation process as was used for the double pendulum task. Clutter parameters that follow are with respect to rendered image size of $500$x$500$px and time step ($dt$) of $0.01$. $50\%$ of clutter is rendered beneath and $50\%$ is rendered on top of the spider. For dynamic clutter, each geometry simulates motion using a random, constant position velocity ($\dot{x}$, $\dot{y}$) and orientation velocity ($\dot{\theta}$). Position velocities are sampled from $\mathcal{N}(0, 3)$ while orientation velocities are sampled from $\mathcal{N}(0, 0.05)$. Clutter is simulated as either a rectangle with $70\%$ probability or a circle with $30\%$ probability. Clutter rectangles are sized randomly with length of $\text{max}(0, l\sim\mathcal{N}(20, 3))$ and height of $\text{max}(0, h\sim\mathcal{N}(80, 5))$. The color of clutter rectangles is chosen uniformly at random from the same colors as were used for the spider arms. Clutter circles are sized randomly with a radius of $\text{max}(0, r\sim\mathcal{N}(10, 3))$ and colored yellow to match the color of the spider's joints. The position of each clutter geometry was randomly initialized within the image boundary.

For the training/validation sequences that included clutter, the number of clutter shapes rendered beneath and on top of the double pendulum was each randomly sampled from independent Binomial distributions using $n=10$, $p=0.5$. The test set was generated with uniformly distributed clutter ratios, as described in \cref{sec:datasets}, using rejection sampling with variable numbers of rendered geometries.

\subsection{Learned Pairwise Inspection}
\label{appendix:pairwise_inspection}

\begin{figure}[b]
\centering
\includegraphics[width=10cm]{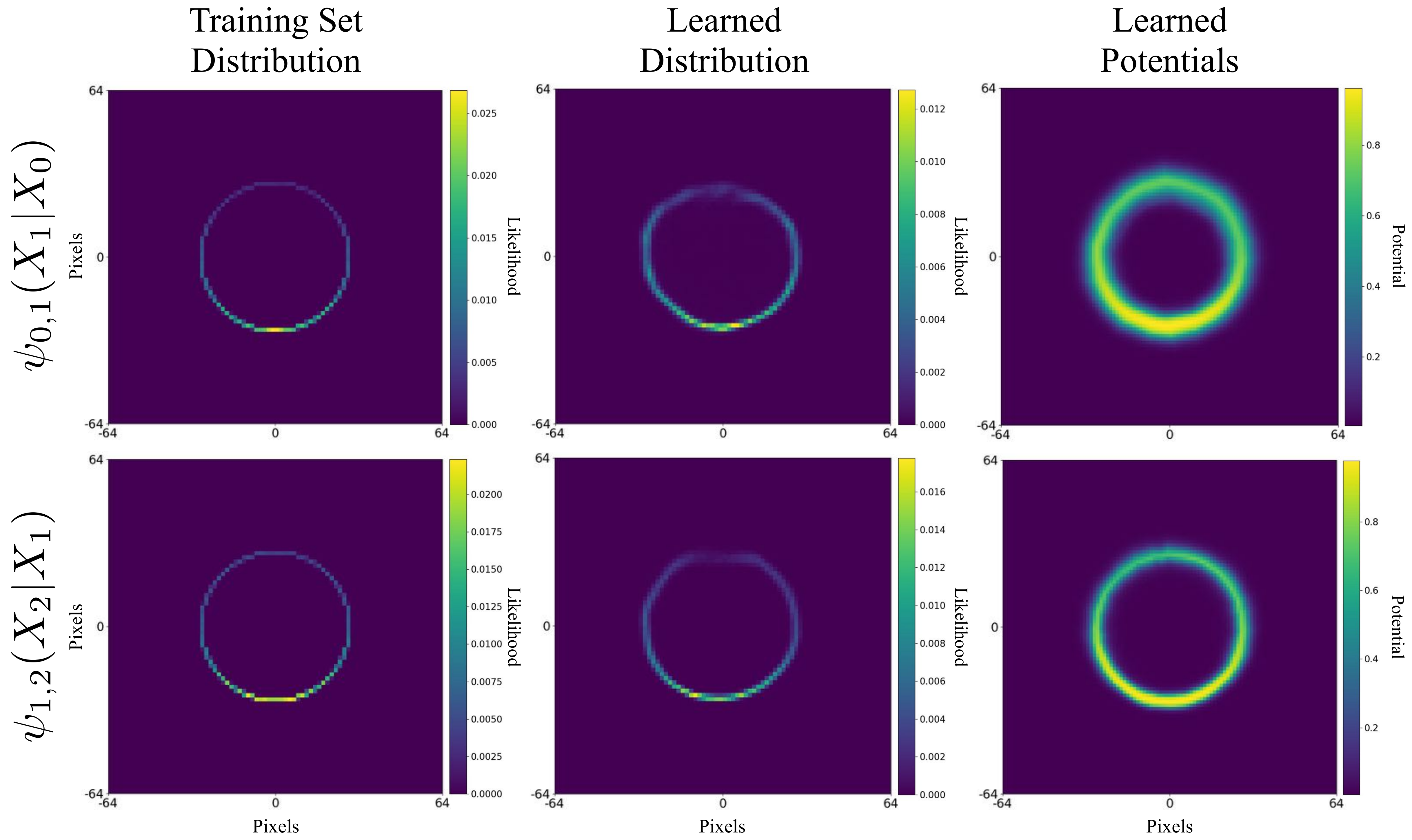}
\caption{\footnotesize Inspection of learned pairwise potentials from double pendulum tracking.}
\label{fig:learned_pairwise_pendulum}
\end{figure}
As further validation of DNBP, the learned pairwise potentials are inspected in \cref{fig:learned_pairwise_pendulum}. The normalized histogram of pairwise translations computed from the training set for $X_1-X_0$ (top) and $X_2-X_1$ (bottom) are shown in the left column of \cref{fig:learned_pairwise_pendulum}. The middle column shows the normalized histogram of samples from learned pairwise sampler networks, $\psi_{0,1}^{\sim}(\cdot)$ and $\psi_{1,2}^{\sim}(\cdot)$. Finally, the right column shows output from the learned pairwise density networks, $\psi_{0,1}^{\rho}(\cdot)$ and $\psi_{1,2}^{\rho}(\cdot)$,  generated with $100$x$100$ uniform samples across pairwise translation space. The qualitative similarity between each learned potential model and the corresponding true distribution of pairwise translations indicates that DNBP is successful in learning to model each pairwise potential factor. The circular pairwise relationships are explained by the fact that each pair of double pendulum keypoints is related by a revolute joint. The effect of simulated gravity in the double pendulum experiment can be observed by the bias of each pairwise potential in favor of the lower half of each plot as indicated by increased likelihood.

\begin{figure}[t]
\centering
\includegraphics[width=10cm]{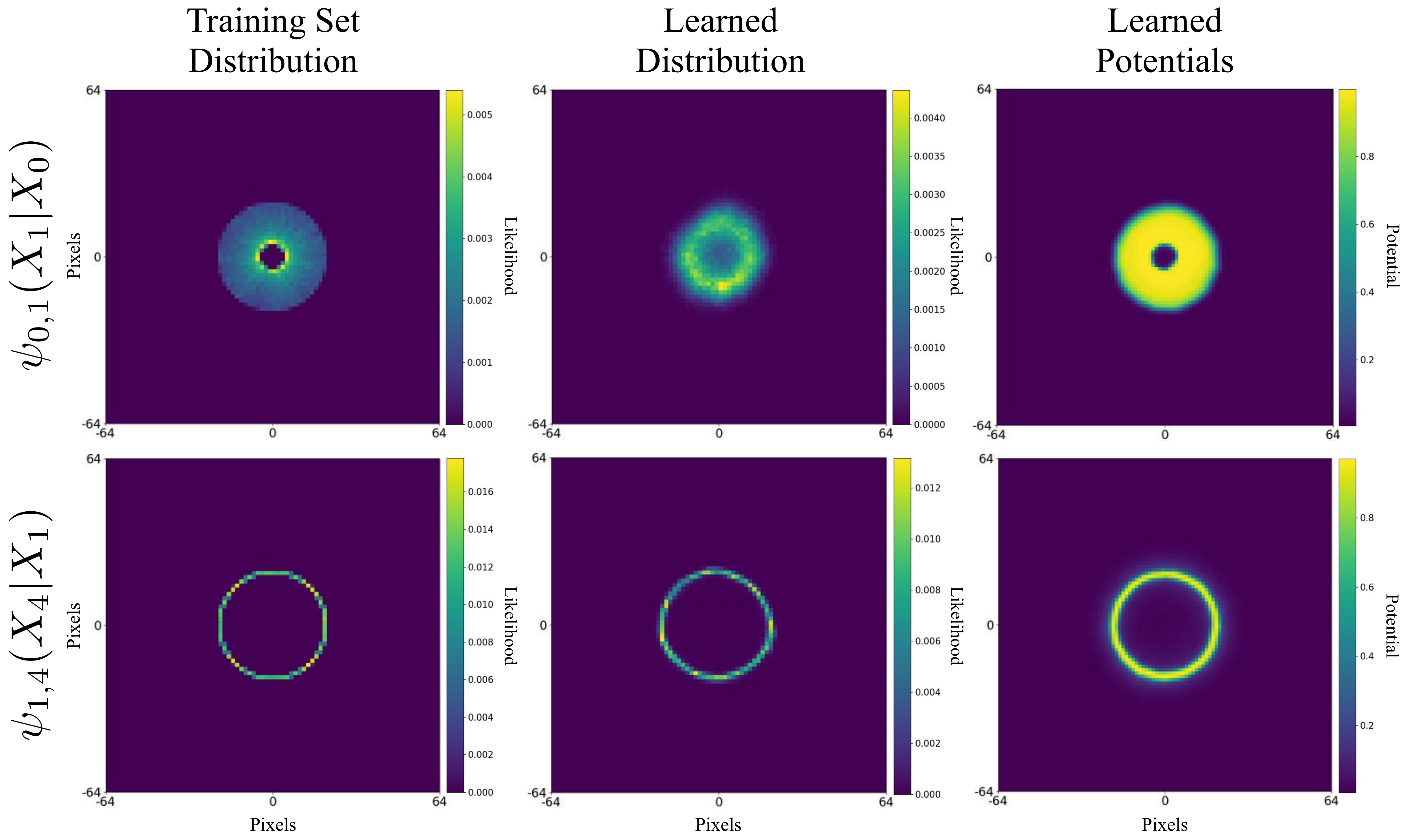}
\caption{\footnotesize Inspection of DNBP's learned pairwise potentials from spider tracking. Only two of the six are shown to avoid redundancy, remaining four show very similar output.}
\label{fig:learned_pairwise_spider}
\end{figure}

The pairwise potential functions learned by DNBP in the spider tracking task are visualized as was done in the double pendulum task. \cref{fig:learned_pairwise_spider} shows qualitative output from two of the six models. Only two are shown to avoid redundancy; chosen results are representative of remaining four potential functions. The left column of \cref{fig:learned_pairwise_spider} shows the normalized histogram of pairwise translations as computed from the training set for $X_1-X_0$ (top) and $X_4-X_1$ (bottom). The middle column of \cref{fig:learned_pairwise_spider} shows the normalized histogram of samples from learned pairwise sampler networks, $\psi_{0,1}^{\sim}(\cdot)$ and $\psi_{1,4}^{\sim}(\cdot)$. Finally in the right column of \cref{fig:learned_pairwise_spider}, uniformly sampled output ($100$x$100$ samples across pixel space) of the learned pairwise density networks is shown. Once again, the visual similarity between each learned potential function and the corresponding true distribution of pairwise translations is an indicator that DNBP is successful in learning to model each pairwise factor. Observe that the learned potential functions for $\psi_{1,4}(\cdot)$, which correspond to a revolute articulation, show no bias in favor of the downward configuration. This result is notably different from the potential functions learned on the double pendulum task and can be explained by the absence of gravity in the spider simulation. Similarly, the learned models for $\psi_{0,1}(\cdot)$ on the spider task exhibit a torus shape due to the effect of prismatic motion associated with the corresponding joint's articulation type and constraint. 

\clearpage
\subsection{Hand Tracking Results}
\label{appendix:hand_tracking_resutls}

\begin{figure}[h]
\centering
\includegraphics[width=0.93\linewidth]{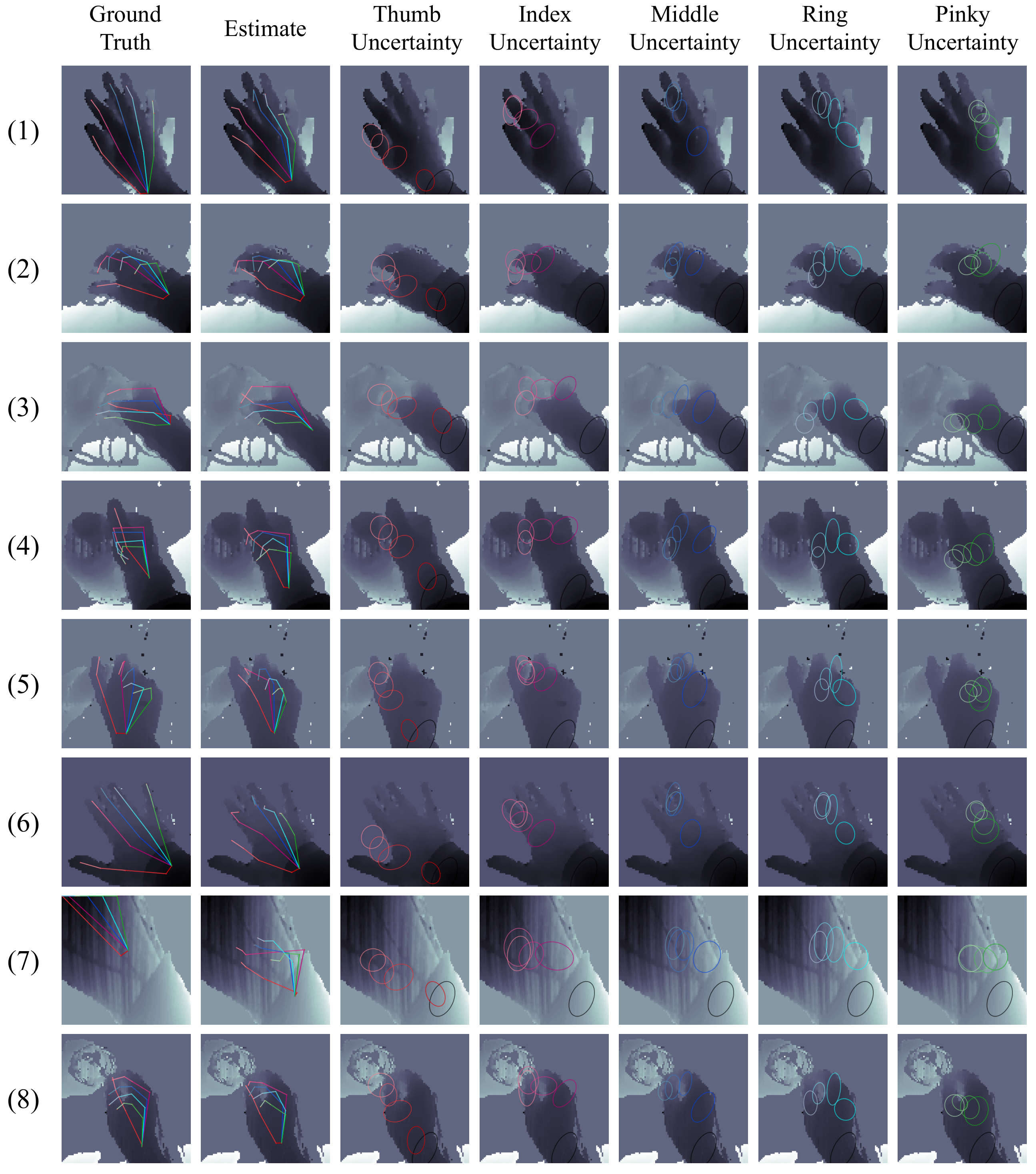}
\caption{\footnotesize Output from DNBP on randomly sampled frames of the hand pose tracking experiment. Visualized model uncertainty is calculated from the marginal belief estimates of DNBP as $1$ standard deviation in the horizontal and vertical dimensions respectively as calculated by estimated covariance of belief particles. Uncertainty in depth dimension is not visualized.}
\label{fig:hand_qual2}
\end{figure}


\end{document}